\documentclass{article}
\usepackage[left=1 in,right=1 in,top=1 in,bottom=1 in]{geometry}

\title{Discordance Minimization-based Imputation Algorithms for Missing Values in Rating Data}

\author{Young Woong Park \thanks{ywpark@iastate.edu; corresponding author} \thanks{Ivy College of Business, Iowa State University, Ames, IA, USA}
\and
  Jinhak Kim\thanks{School of Business Administration, Ajou University, Suwon, Republic of Korea}
   \and
  Dan Zhu\footnotemark[2] 
  }

\usepackage{verbatim} 

\date{\today}
\usepackage[pdftex]{graphicx}

\usepackage[round]{natbib}
\usepackage{graphicx}
\usepackage{amsmath,amsthm,amssymb}
\usepackage{enumitem} 
\usepackage{algorithmic}	
\usepackage{algorithm}
\usepackage{url} 
\usepackage[hidelinks]{hyperref}
\usepackage{xcolor}
\usepackage{colortbl} 
\usepackage{subfigure}
\usepackage{titling} 
\usepackage{tikz}
\usepackage{svg}
\usepackage{multirow}

\usepackage{diagbox}

\newcommand*{\shifttext}[2]{%
  \settowidth{\@tempdima}{#2}%
  \makebox[\@tempdima]{\hspace*{#1}#2}%
}

\def\O{\mathcal{O}}

\def\E{\mathcal{E}}

\theoremstyle{definition} \newtheorem*{example}{Example}
\newtheorem{definition}{Definition}
\newtheorem{assumption}{Assumption}

\newtheorem{theorem}{Theorem}
\newtheorem{corollary}{Corollary}
\newtheorem{lemma}{Lemma}
\usepackage{etoolbox}

\begin{document}

\maketitle
\vspace{-0.5cm}

\begin{abstract}
Ratings are frequently used to evaluate and compare subjects in various applications, from education to healthcare, because ratings provide succinct yet credible measures for comparing subjects. However, when multiple rating lists are combined or considered together, subjects often have missing ratings, because most rating lists do not rate every subject in the combined list. In this study, we propose analyses on missing value patterns using six real-world data sets in various applications, as well as the conditions for applicability of imputation algorithms. Based on the special structures and properties derived from the analyses, we propose optimization models and algorithms that minimize the total rating discordance across rating providers to impute missing ratings in the combined rating lists, using only the known rating information. The total rating discordance is defined as the sum of the pairwise discordance metric, which can be written as a quadratic function. Computational experiments based on real-world and synthetic rating data sets show that the proposed methods outperform the state-of-the-art general imputation methods in the literature in terms of imputation accuracy.
\end{abstract}

\section{Introduction}

Ratings are ubiquitous: university and school ratings for students; credit ratings and environmental, social, and governance ratings for investors; hospital ratings for patients; journal ratings for researchers and research institutions; and so on. Despite the criticism that a single number or grade cannot characterize the performance of the subjects evaluated, ratings are used in everyday individual and managerial decision-making processes because they usually provide succinct yet credible performance measures and references. Due to their importance, a substantial body of research has been conducted on the impacts of ratings (or rankings) in various contexts, such as the impact of sovereign credit ratings on markets \citep{cantor1996determinants}, the impact of product rankings on consumer behavior \citep{ghose2014examining}, the effects of online hotel ratings on customers' booking intentions and behavior toward a hotel \citep{casalo2015online}, the impact of college rankings and their visibility on students’ application decisions \citep{luca2013salience}, the effect of school ratings on neighborhood choice and home values \citep{Lerner}, the impact of ranking systems on the decision-making of higher education institutions \citep{hazelkorn2007impact}, and the effects of hospital performance ratings on the public \citep{hibbard2005hospital}.

Although ratings have a significant impact, a single rating system does not evaluate or rate all assessable subjects. Many rating systems rate a limited number of subjects because of capacity limitations in collecting data or performing analysis, insufficient data for evaluations, and the lack of interest in specific groups of subjects. 
To compare the known ratings of all subjects, a \textit{combined list} can be created by concatenating the ratings of multiple rating systems. However, in the tabular form of the combined list, missing entries are frequently observed because a subject rated by one rating system is not rated by another.
The missingness of the combined list can be problematic, considering the impact of ratings on decision-making. For example, journal rating lists are frequently used as credible measures for evaluating the quality of faculty scholarship \citep{kim2019}. However, a combined journal rating list includes a significant number of missing entries for many reasons, such as quality control at rating agencies and scope of the rating. The missingness often creates challenges in faculty scholarship evaluation processes, particularly when the academic institution adopts a single rating list that does not rate some of the institutions' target journals.

To remedy this issue, traditional research has primarily focused on reverse-engineering the underlying rating formula or identifying significant explanatory factors that constitute the rating system \citep{chang2012reverse, adelman2020efficient}.
However, the applicability of the traditional approaches is limited for imputing missing values on a combined list of multiple rating systems, because the explanatory factors may vary across rating systems, thus requiring customized treatment for each rating system.
To overcome this limitation, \citet{kim2019} recently proposed imputing missing entries in a combined rating list, based solely on \emph{data imputation} approaches.

\begin{figure}[ht]
\begin{center}
    \subfigure[Character-coded ordinal rating matrix]{%
           \includegraphics[scale=0.55]{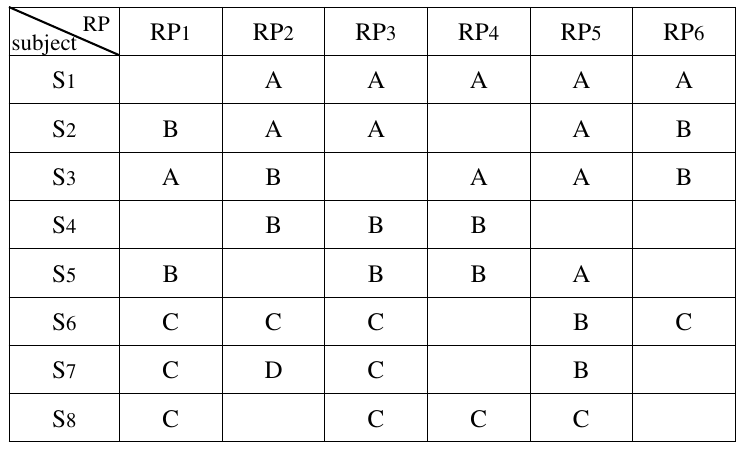} \label{fg_example_matrix_categorical}
        }\qquad \quad
        \subfigure[Numerical ordinal rating matrix]{%
          \includegraphics[scale=0.55]{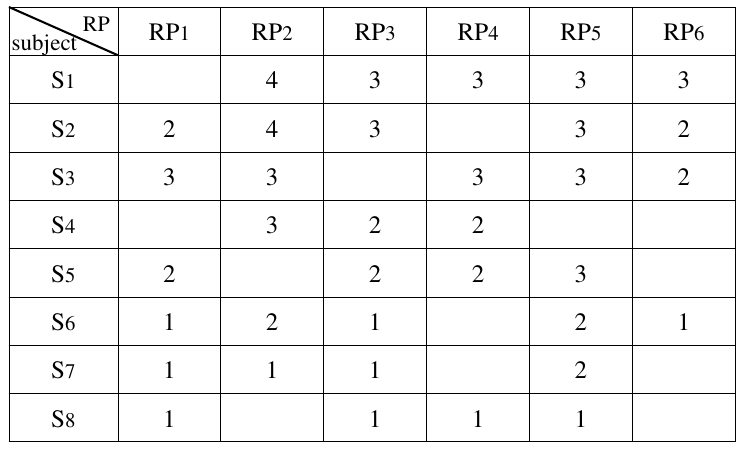} \label{fg_example_matrix_ordinal}
    }
\end{center}
\vspace{-0.3cm}
    \caption{Example rating matrices}
\label{fg_example_matrix}
\end{figure}

The goal of this research is to integrate multiple rating systems from various sources, create combined rating lists, and impute the missing values of the combined matrix using only the known ratings. The resulting imputed rating matrix will enable users to access all ratings by all rating providers in their interests. We study the missing value imputation problem for a matrix of combined rating lists with a structure similar to the one in Figure~\ref{fg_example_matrix}, 
which essentially shares the same objective with matrix completion and collaborative filtering.
We assume that the ratings must be ordinal. To quantify our model, any character-coded ordinal ratings are converted to numerical ordinal ratings whose categories are natural numbers.
For example, the character-coded ordinal ratings in Figure~\ref{fg_example_matrix_categorical} are converted into numerical ordinal ratings in Figure~\ref{fg_example_matrix_ordinal}. Without loss of generality, we assume the higher rating values reflect superior quality. In the example data matrices Figure~\ref{fg_example_matrix}, six rating providers (RPs) rate subsets of the subjects: RP1 rates six subjects, RP2 rates six subjects, etc.

Data imputation has been widely studied and has a rich literature. The problem we consider belongs to the general data imputation problem, but it requires ordinal data values and prefers coherent columns. Hence, we next provide a brief overview of frequently used data imputation techniques discussed in the literature. The data imputation methods can be classified into two categories \citep{garcia2010pattern,lin2020missing}: statistical methods and machine learning-based methods.
In particular, \cite{lin2020missing} list a few of the most widely used techniques in the literature in each of these categories. In the next two paragraphs, we briefly review the methods in statistical and machine learning categories.

In the statistical method category, expectation-maximization (EM), regression-based methods, and mean/mode imputation are the most commonly used subcategories.
EM is widely used in various applications and is an iterative method in which each iteration consists of two steps: (1) E-step updates the conditional expectation of the log likelihood given the observed data and the current parameters, and (2) M-step finds the new parameter set by maximizing 
the conditional expectation of the log likelihood. See Chapter 5 of \cite{schafer1997analysis} for more detail. In R, multiple packages offer EM-based imputation methods including \emph{norm, imputeR}, and \emph{TestDataImputation}.
A regression-based method sets a variable that includes missing entries as a response variable, while setting other variables as predictor variables to build a linear or logistic multiple regression model as a predictive model. There are several variants of regression-based imputation models implemented in R. For example, the \emph{VIM} \citep{kowarik2016imputation} package includes a regression-based imputation function \emph{regressionImp}.
In practice, a simple mean/mode imputation method is popular because of its simplicity and computational efficiency. A mean (resp. mode) imputation replaces each missing entry with the mean (resp. mode) of the row (or column) that includes the missing entry.
In addition, multiple imputation methods have recently received greater attention from researchers. Multiple imputation methods generate multiple imputed data sets, where each imputed data set can be used for the same analysis, and all results can be used to conclude. When a single imputed data set is needed, we can aggregate from the multiple imputed data sets. Multiple imputation by chained equation (MICE) \citep{buuren2011mice} is a representative multiple imputation algorithm in the R package \emph{mice}.

In the machine learning methods category, the four most widely used imputation methods are sub-categorized into clustering-based methods, decision tree-based methods, $k$-Nearest Neighbors methods, and random forest-based methods.
Clustering-based methods first place the observations into several clusters, where the number of clusters is user-defined, and the distances between observations are measured directly from the incomplete data. Then, the missing entries in the same cluster are imputed based only on information within the cluster. For example, nearest neighborhood methods \citep{patil2010missing} can be used to determine the nearest instance within the cluster, which then determines the imputation.
Some clustering-based imputation methods are implemented in the R package \emph{ClustImpute}.
Decision tree-based methods impute the missing entries in a variable, using a decision tree as a predictive model. The decision tree is constructed based on the remaining variables (other than the imputed one). Classification and Regression Trees (CART) \citep{breiman2017classification} is a widely used decision tree-based method implemented in the \emph{mice} package in R.
Random forest-based methods are based on multiple decision trees constructed using bootstrapping. The imputation is performed by aggregating the predictions from the decision trees.
A fast random forest-based imputation method is implemented in the well-known R package \emph{missForest} \citep{stekhoven2012missforest}.

In addition to statistical and machine learning methods, optimization-based methods have also been studied for data imputation. In the seminal work by \cite{candes2009exact}, the authors propose a convex optimization formulation for matrix completion, which finds the imputed matrix that minimizes the nuclear norm.
Motivated by this convex formulation, an alternative convex formulation has been proposed by \cite{mazumder2010spectral}, by which the method \emph{softImpute} was first implemented.
The \emph{softimpute} algorithm is an iterative method that, in each iteration, uses a  soft-thresholded singular value decomposition to impute the missing values of the filled-in matrix from the previous iteration.
\cite{hastie2015matrix} improved the previous version of \emph{softImpute} based on the enhanced matrix factorization algorithm.
Several variants of these methods and other computational algorithms have been proposed over the past decade. For example, see \cite{ramlatchan2018survey} for a review of these methods.
More recently, \cite{kim2019} have proposed a mixed-integer linear programming formulation for data with multiple ordinal variables, which can be solved using commercial optimization solvers.

Our contributions are threefold. 
\begin{enumerate}[noitemsep]
    \item We introduce various real-world rating data sets and study the properties of the input rating data matrix and missing value patterns. First, Mann–Whitney U test is used to test if the missing value implies inferiority or superiority. Second, the Kendall rank correlation is used to measure the consensus levels among the rating providers. These two analyses are performed for multiple real-world rating data from various applications. We also provide synthetic data and a data-generation procedure (based on the properties we observed from the real data) for systematic performance comparison.
    \item Based on the properties and structure of the rating data analyzed, we propose a quadratic programming (QP) formulation, which minimizes the total discordance between RPs in the combined rating lists. In contrast to the existing studies, our algorithms weight each pair of RPs based on their consensus levels, where the weights emphasize highly correlated RP pairs. To improve the scalability of the proposed QP model, we propose a decomposable version of the QP in which the imputation of each cell is independent of the other. We derive closed-form solutions to both the QP and its decomposable variant. This enables us to implement scalable imputation algorithms that do not depend on commercial solvers but only require solving systems of linear equations. Finally, we propose a mathematical procedure and definitions, which test if the proposed algorithms are well-defined and applicable. The derived conditions are referred to as estimatability and level-1 estimatability.
    \item The computational experiment, based on real-world and synthetic data sets, shows that the analytical solution approaches significantly reduce the time and space complexities of the proposed QP model (solved using a commercial solver); they also outperform the state-of-the-art benchmark algorithms reported in the literature for imputing missing ratings.
\end{enumerate}

The paper is structured as follows. In Section \ref{section_analysis}, we present real-world data sets and conduct analyses on the properties of the input rating data matrix and missing value patterns. In Section \ref{section_model_algorithm}, mathematical models and algorithms are proposed and sufficient conditions for the algorithms are derived. In Section \ref{section_experiment}, we compare the proposed algorithms' performance with popular imputation algorithms in the literature using real-world and synthetic data.

\section{Rating Data: Analysis and Imputed Data Usage}
\label{section_analysis}

In this section, we present and analyze multiple real-world data sets from various applications. First, in Section \ref{subsection_real_data}, we briefly describe the real-world data sets we compiled. Next, we focus on checking whether the missing values imply inferiority or superiority in Section \ref{subsec_analysis_missing_value_patterns}. We propose a procedure to measure the consensus level between a pair of RPs in Section \ref{subsec_analysis_consensus_level}. The consensus levels we define are used to penalize the effects of discordant pairs of rating providers on the overall discordance. Finally, in Section \ref{subsection_imputed_data_usage}, we exemplify how users can use the resulting imputed rating data.
In the rest of the paper, we use the notations summarized below.
\begin{enumerate}[noitemsep]
    \item[] $X$: $m \times n$ rating matrix with missing entries
    \item[] $I = \{1,\cdots,m\}$: index set of rows (subjects) of matrix $X$
    \item[] $J = \{1,\cdots,n\}$: index set of columns (rating providers) of matrix $X$
    \item[] $x_{ij}$: entry at row $i \in I$ and column $j \in J$ of matrix $X$
    \item[] $u_j = \max_{i \in I} \{ x_{ij}\}$: maximum rating value for rating provider $j \in J$
    \item[] $l_j = \min_{i \in I} \{ x_{ij}\}$: minimum rating value for rating provider $j \in J$
    \item[] $c_j = u_j - l_j + 1$: number of observable ratings categories for rating provider $j \in J$
    \item[] $S = \{(i,j) \in I\times J \}$: index set of matrix entries (Cartesian product of $I$ and $J$)
    \item[] $S^{kl} = \{ (i,j) \in (I\setminus\{k\})\times (J\setminus\{l\}) \}$: index set of matrix entries excluding row $k$ and column $l$
    \item[] $Q \subset S$: index set of missing entries
    \item[] $\Phi = \{ x_{kl} | (k,l) \in Q \}$: missing values to impute (set of decision variables)
    \item[] $\hat{X}$: an imputed matrix, where $\hat{x}_{ij}$ is the entry at row $i \in I$ and column $j \in J$
    \item[] $w_{lj}$: weight imposed for rating provider pairs $l \in J$ and $j \in J$
\end{enumerate}

\subsection{Real-World Rating Data}
\label{subsection_real_data}

Before we present the in-depth analysis in the later sections, we introduce six real-world rating data sets we complied. 
The data sets used in our analyses and experiments are from various applications: hospital rating, journal rating, environmental, social, and governance (ESG) rating, elementary and high school rating, and movie rating. The rating values in these applications offer credible measures to evaluate organizations, individuals, and subjects. However, when multiple RPs are considered together, many missing values exist. This hurts the usefulness of the rating systems, and accurate imputation of the missing ratings becomes important \citep{kim2019}.
The detailed background and data collection procedures, conversion procedures\footnote{Some RPs use continuous scores, and we convert them into ordinal ratings.}, and missing rate distributions are described in Appendix A1-A5, A6, and A7, respectively.

In Table \ref{table_data_summary}, the summary statistics of the six data sets are presented. The second and third columns present the numbers of subjects rated and the numbers of RPs, respectively. The next four columns include the average, median, minimum, and maximum missing rates of the RPs (columns). The last two columns show the percentages of the rows with only one rating and all ratings, respectively. The column summary statistics indicate that some RPs rate nearly all subjects (e.g., an RP in the Journal data set has missing values for 3.8\% of the journals), while some RPs rate very few subjects (e.g., an RP for the US Hospital data set has missing values for 94.5\% hospitals). The row summary statistics indicate that some data sets have near-zero rows that have only one rating available (e.g., the journal data set has 0.4\% of the rows with one rating). In comparison, some data sets have more than half of rows with only one rating (e.g., the High School data set has 55.3\% of the rows with one rating). 
See Appendix A7 for more detailed statistics and distributions of the missing rates.

\begin{table}[htbp]
  \centering
  \begin{small}
    \begin{tabular}{|c|c|c|cccc|cc|}
    \hline
    \multirow{2}{*}{Data} & \multirow{2}{*}{\# Rows} & \multirow{2}{*}{\# Cols} & \multicolumn{4}{c|}{Column Missing Rates} & \multicolumn{2}{c|}{\% Rows Rated} \\
          &       &       & Avg & Med & Min & Max & One RP & All RPs \\ \hline
    US Hospital & 4217  & 4     & 30.2\% & 26.9\% & 11.6\% & 55.2\% & 21.0\% & 41.3\% \\
    Journal & 944   & 11    & 36.0\% & 37.7\% & 5.2\% & 65.8\% & 1.2\% & 13.5\% \\
    ESG & 1356  & 4     & 47.5\% & 53.9\% & 3.8\% & 78.4\% & 44.0\% & 14.0\% \\
    Elementary School & 301   & 5     & 38.9\% & 33.9\% & 25.9\% & 66.8\% & 29.9\% & 24.3\% \\
    High School & 132   & 5     & 55.6\% & 53.8\% & 28.8\% & 81.8\% & 55.3\% & 11.4\%  \\
  Movielens & 1102 & 12 & 62.7\%	& 64.8\% & 53.0\% & 67.3\% & 23.1\% & 3.0\%\\
    \hline
    \end{tabular}%
\end{small}
      \caption{Summary statistics of data sets}
  \label{table_data_summary}%
\end{table}%

\subsection{Missingness and Inferiority}
\label{subsec_analysis_missing_value_patterns}
Most multiple imputation algorithms assume a missing data mechanism, called \emph{missing at random (MAR)} \citep{rubin1976inference}, although this assumption is frequently violated. 
In particular, we focus primarily on rating data in which the missing at random condition is easily violated because the probability that a rating is missing does depend on the value of that rating \citep{marlin2009collaborative}.
As partial evidence of this statement,
we check if the missingness has some associations with inferiority. 
More specifically, for each RP pair $(j,l)$, we use the Mann–Whitney U test\footnote[1]{Mann–Whitney U test is a nonparametric test that, for two groups G1 and G2, checks if the probability of G1 being greater than G2 is equal to the probability of G2 being greater than G1. It only requires that you are able to rank order the individual scores or values; there is no need to compute means or variances.} to compare the medians between two groups of RP $l$'s ratings: (i) group $G_{jl}^1$ includes RP $l$'s ratings for the subjects whose ratings of RP $j$ are missing and (ii) group $G_{jl}^2$ includes RP $l$'s ratings for the subjects whose ratings of RP $j$ are available. For RP pair $(j,l)$, we test the following hypotheses.
\begin{enumerate}[noitemsep]
    \item[] $H_0$: $P[ x' > x'' | x' \in G_{jl}^1, x'' \in G_{jl}^2] = 0.5$ 
    \item[] $H_1$: $P[ x' > x'' | x' \in G_{jl}^1, x'' \in G_{jl}^2] \neq 0.5$ 
\end{enumerate}

\begin{figure}[ht]
     \begin{center}
        \subfigure[Hospital]{%
          \includegraphics[scale=0.55]{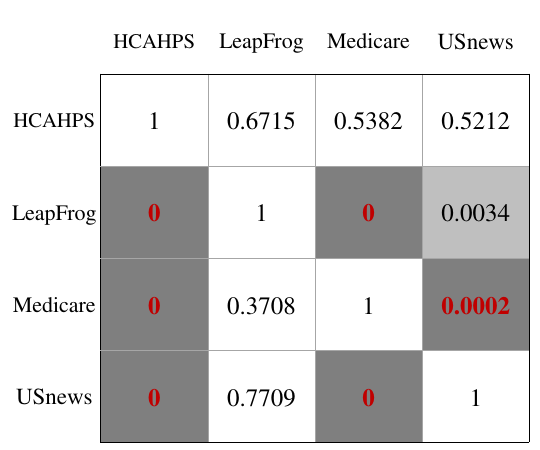} \label{fg_hospitals_pval}
        }\quad
        \subfigure[Journal]{%
          \includegraphics[scale=0.55]{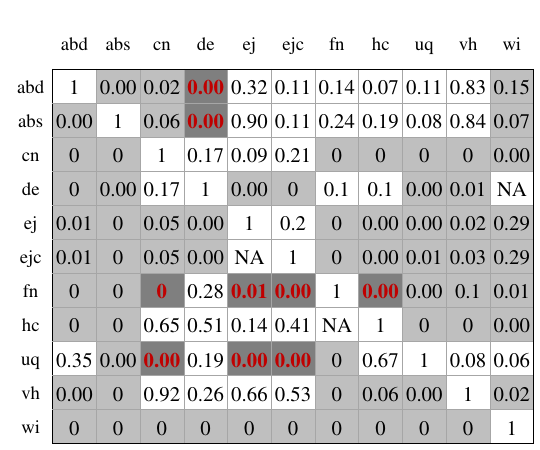} \label{fg_journal_pval}
        }\quad
        \subfigure[ESG]{%
          \includegraphics[scale=0.55]{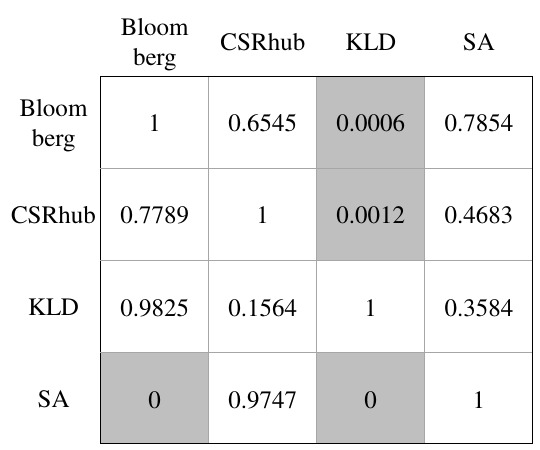} \label{fg_sustainability_pval}
        }\quad
        \subfigure[Elementary School]{%
          \includegraphics[scale=0.55]{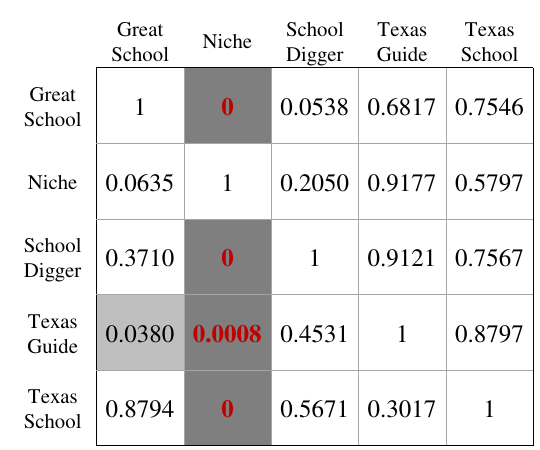} \label{fg_school_elementary_pval}
        }\quad
        \subfigure[High School]{%
          \includegraphics[scale=0.55]{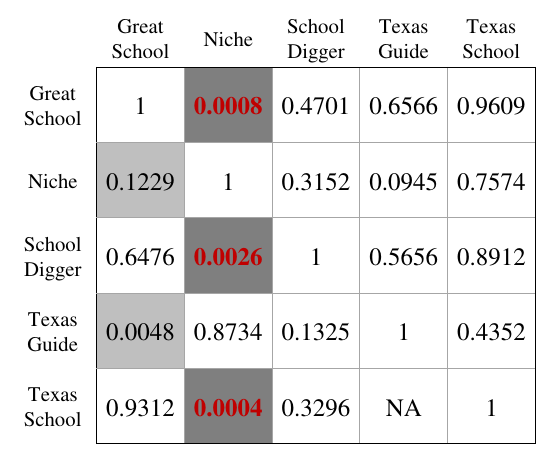} \label{fg_school_high_pval}
        }\quad
        \subfigure[Movielens]{%
          \includegraphics[scale=0.55]{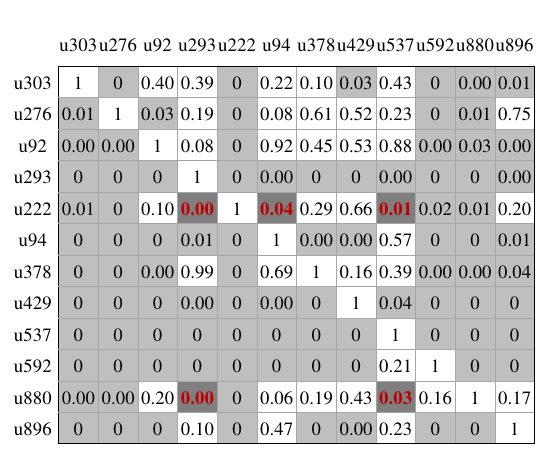} \label{fg_movielens_pval}
        }
    \end{center}
    \vspace{-0.3cm}
    \caption{p-value matrices from Mann–Whitney U test}
\label{fg_pval}
\end{figure}

If the U test concludes that the two groups are different, then missing ratings in RP $l$ may imply the inferiority or superiority of the subjects. In Figure \ref{fg_pval}, each matrix reports the p-values of the U test of RP pairs. For each row $j$ and column $l$ of each matrix, the matrix entry represents the p-value of the test for RP pair $(j,l)$ that compares the average RP $l$ ratings of two subject groups, where the first group consists of subjects unrated by RP $j$ and the second group consists of subjects rated by RP $j$. For example, in Figure \ref{fg_hospitals_pval}, in Row 1 and Column 2, 0.6715 is the p-value of the test that compares the ratings of LeapFrog with and without the missing values of HCAHPS. The NA cells in Figure \ref{fg_journal_pval} and \ref{fg_school_high_pval} indicate that the test is not available for the corresponding pair. The test result indicates that there is no significant difference in LeapFrog ratings between the two groups, with or without the missing values of HCAHPS. In all matrices, the cells are highlighted with light gray color and black font if the p-value is less than 0.05 and the missing ratings in RP $j$ mean worse ratings in RP $l$. If the p-value is less than 0.05 and the missing ratings in RP $j$ mean better ratings in RP $l$, then the number is colored in red in darker gray cells. The color-coded matrices show that missing values generally indicate inferiority for Journal and Movielens data sets. In contrast, it is not sufficiently clear to conclude similar results for the Hospital, ESG, Elementary School, and High School data sets. For the Elementary and High School data sets, missing values generally mean better ratings for Niche.

\subsection{Analysis for Consensus Level}
\label{subsec_analysis_consensus_level}
Our imputation methods incorporate the discordance levels for all pairs of RPs and aim to minimize their sum.
Then, each discordance level is penalized based on the consensus level obtained from the observed entries.
To measure the consensus level for each pair of RPs, we use the Kendall rank correlation coefficient $\tau_{B}$, which is one of the most widely used distance measures between two ranked lists in the rank aggregation literature. For example, see \cite{lin2010rank}, \cite{dwork2001rank}, and \cite{fagin2003efficient} for your reference.
Unlike the typical Pearson correlation, the Kendall rank correlation focuses on the order between the two pairs, which aligns with our approach of minimizing the discordance. 

\begin{figure}[ht]
     \begin{center}
        \subfigure[Hospital]{%
          \includegraphics[scale=0.55]{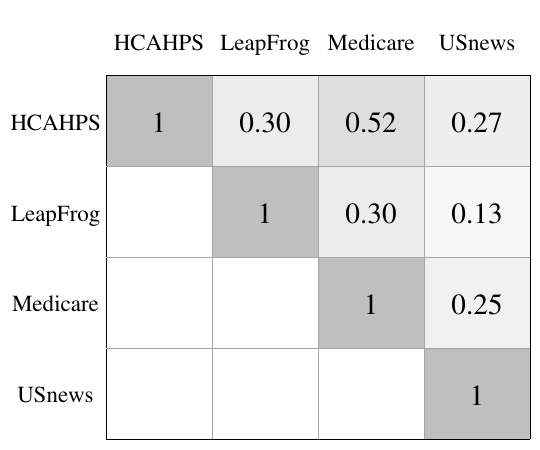} \label{fg_hospitals_kendall}
        }\quad
        \subfigure[Journal]{%
          \includegraphics[scale=0.55]{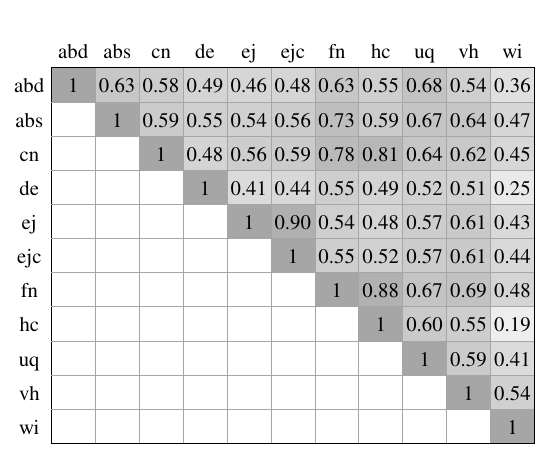} \label{fg_journal_kendall}
        }\quad
        \subfigure[ESG]{%
          \includegraphics[scale=0.55]{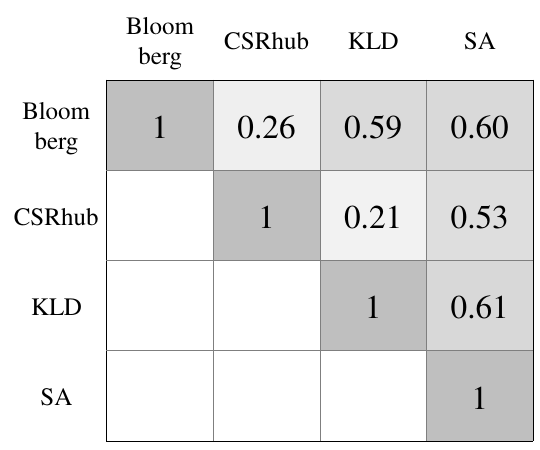} \label{fg_sustainability_kendall}
        }\quad
        \subfigure[Elementary School]{%
          \includegraphics[scale=0.55]{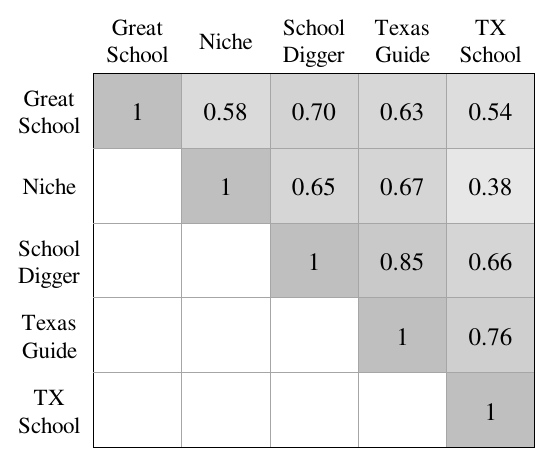} \label{fg_school_elementary_kendall}
        }\quad
        \subfigure[High School]{%
          \includegraphics[scale=0.55]{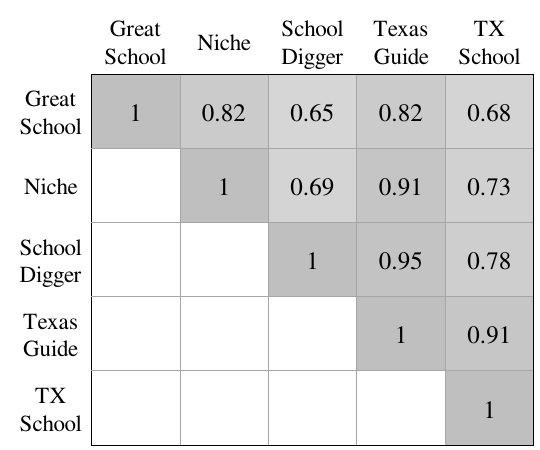} \label{fg_school_high_kendall}
        }\quad
        \subfigure[Movielens]{%
          \includegraphics[scale=0.55]{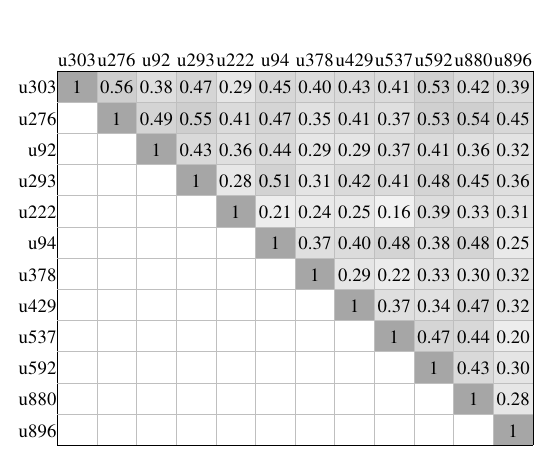} \label{fg_movielens_kendall}
        }
    \end{center}
    \vspace{-0.3cm}
    \caption{Kendall rank correlations}
\label{fg_kendall}
\end{figure}

Figure \ref{fg_kendall} presents the correlation matrices of the six data sets. The darker and lighter cells have higher and lower correlations, respectively. For all of the RP pairs in the data sets, the correlation coefficients are positive, indicating that the ratings are positively correlated. However, the magnitudes of the correlations vary for different RP pairs and data sets. The correlations for the Journal data range from 0.19 to 0.90, with a mean of 0.56; correlations for the Hospital data range from 0.13 to 0.52 with a mean of 0.29. The average correlation coefficients for the Elementary School, High School, and ESG data sets are 0.64, 0.79, and 0.47, respectively. Hence, we can see that school ratings are more consistent across different RPs, while the hospital rating data set has low consensus levels among different RPs. This observation suggests that highly correlated RP pairs can refer to each other for imputing their missing values. Further, highly correlated pairs should affect each other more, and lowly correlated pairs should affect each other less. In Section \ref{section_model_algorithm}, we propose using these Kendall rank correlation matrices to weight the RP pairs.

\subsection{Imputed Rating Data Usage}
\label{subsection_imputed_data_usage}

Data imputation serves as a pre-analysis tool in many imputation tasks to prepare a complete data matrix for the main analysis, such as regression analysis, by filling in missing values. 
On the other hand, the direct utilization of data imputation is becoming increasingly important, as illustrated in Section~\ref{subsection_real_data}.
If an end-user prefers specific RPs and the subjects of interest are not rated by the preferred RPs, an accurate estimation of the missing values can assist stakeholders in making fair and reasonable decisions.
For instance, numerous academic institutions use a journal quality list for evaluating scholarships, relying on the ratings provided by this list as a quality score for a published journal \citep{kim2019}. However, high-quality journals are often excluded from such trusted journal quality lists. In such cases, imputed ratings can provide quality scores for the missing journals in the designated journal quality list. Similarly, imputed ratings for hospitals, companies, and schools can provide valuable information and insight to users.

An analyst interested in estimating the missing ratings of a preferred rating list can use the proposed algorithm as follows. First, generate the Kendall rank correlation matrix using all available data. Second, select a subset of RPs (columns) relevant or highly correlated to the designated RP. Third, use the rating data of the selected RPs to run the proposed algorithm. Finally, the imputed ratings can be used to evaluate subjects with missing ratings. For example, consider the elementary school data in Table \ref{table_data_summary}. Suppose the analyst is interested in imputing the ratings for GreatSchools. With a threshold of 0.6, the Kendall correlation matrix in Figure \ref{fg_school_elementary_kendall} indicates GreatSchool, SchoolDigger, and Texas Guide should be used for the algorithm. Using the selected RPs and their ratings, the proposed algorithm imputes the missing ratings in the three selected RPs. The analyst can now use the imputed ratings for GreatSchool.

\section{Quadratic Programming-based Imputation Models and Algorithms}
\label{section_model_algorithm}

Missing rating imputation problems have multiple characteristics that make them difficult to solve. First, there are inconsistencies among the ratings, as illustrated in Section \ref{subsec_analysis_consensus_level}. Ideally, when experts have the same opinions on subjects, each subject would get the same rating by all RPs. However, for various reasons, ratings by the RPs for the same subject may differ. For example, in Figure~\ref{fg_example_matrix_ordinal}, RP1 rates S3 higher than S2, whereas RP2 rates S2 higher than S3. This is referred to as an \textit{upset} \citep{kim2019}. Second, rating distributions are not consistent across the RPs. For example, in the Journal rating data set, both Ejis2007 and ABDC2019 use four-category rating scales. However, the proportions of the ratings in the four categories are 14\%, 31\%, 34\%, and 21\% in Ejis2017, while they are 20\%, 48\%, 26\%, and 6\% in ABDC2019. This can be problematic if a simple approach, such as imputation by averaging the available ratings, is used.
Third, the MAR mechanism, which several imputation algorithms assume, can be violated. For example, missingness might be significantly affected by the inferiority of subjects, as discussed in Section \ref{subsec_analysis_missing_value_patterns}.
In this section, we first derive sufficient conditions for our algorithms and then propose quadratic programming-based models and algorithms to overcome these challenges.

\subsection{Sufficient Conditions for Proposed Imputation Algorithms}
\label{subsec_data_assumption}

In this section, we discuss the estimatability of the data and the sufficient conditions for the proposed algorithms to work properly. In other words, we derive sufficient conditions for the data matrix for our algorithms to be well-defined. First, we make the following assumption in the remainder of the paper.

\begin{assumption} 
Every subject (row) is rated by at least one rating provider (column). \label{assumption_rate_by_one}
\end{assumption}
Any imputation algorithm will fail (or return meaningless random output) if Assumption \ref{assumption_rate_by_one} does not hold. For this reason, if the data set does not satisfy Assumption 1, empty rows must be deleted before running the algorithms. Our algorithm, proposed in Section~\ref{section_model_algorithm}, requires a stronger assumption in order to be well-defined and work properly. In this section, we define a condition called \emph{estimatability} that the input data must satisfy for our algorithm. 

We first illustrate the meaning of this notion of estimatability. 
In this paper, we consistently use the index $i$ (and its variations such as $i', i'', i_1, i_2$, etc.) to denote rated subjects, and similarly, the index $j$ (and its variations such as $j', j'', j_1, j_2$, etc.) to represent rating providers, in accordance with the notations introduced in Section~\ref{section_analysis}.
Consider imputing one missing entry associated with subject $i$ and rating provider $j$. Our imputation models explore every entry $(i', j')$ with $i'\ne i$ and $j'\ne j$ and the associated corner entries $(i, j')$ and $(i', j)$; they check discrepancy between ratings for $i$ and $i'$ by $j$ and $j'$.
Notice that the entries $(i, j), (i',j'), (i',j)$, and $(i, j')$ form a $2\times 2$ submatrix of $X$. 
In order for $(i, j)$ to be ``estimatable," we require the existence of $(i', j')$ such that the entries $(i',j'), (i, j')$, and $(i', j)$ are observed or estimatable.
In other words, two rating providers, $j$ and $j'$, can directly share their rating information to estimate unobserved ratings if there is at least one subject that is rated by both rating providers. 
Even when there is no such subject, they can indirectly share their rating information if there is another rating provider $j''$ that mediates them. That is, there exist subjects $i$ and $i'$ such that $i$ rated by both $j$ and $j''$ and $i'$ is rated by both $j''$ and $j'$.
If all rating providers are connected in this manner, we say that the data is estimatable.
Therefore, we define the estimatability of a data entry inductively, where initial estimatability is defined by the existence of $(i', j')$ such that the three corner entries are all ``observed." 

We next formalize the notion of the level-$v$ estimatability with the following definitions.

\begin{definition}
\label{def:entryestimatability}
An unobserved data entry $(i, j)$ is called \emph{level-1 estimatable} if there exists $(i', j')$ such that $i' \ne i$, $j' \ne j$, and all three entries $(i',j), (i, j')$, and $(i',j')$ are observed.
An unobserved entry $(i, j)$ is called \emph{level-$v$ estimatable} if both of the following two conditions hold: (i) It is not level-$u$ estimatable for all $u \in \{1,\dots,v-1\}$; (ii) There exists $(i', j')$ such that $i' \ne i$, $j' \ne j$, and that each of the three entries with indices $(i',j), (i, j')$, and $(i',j')$ is either observed or level-$u$ estimatable for some $u\in\{1,\dots,v-1\}$.
\end{definition}

\begin{definition}
\label{def:dataestimatability}
An entry of the input data $X$ is called \emph{estimatable} if it is level-$v$ estimatable for some positive integer $v$.
\end{definition}

We next define a data set $X$ being estimatable. To this end, we introduce some notations.
Let $E_0$ be the set of all the observed entries of the input data set $X$.
We denote by $E_v$ the set of all level-$v$ estimatable entries of $X$.

\begin{definition}
\begin{enumerate}
\item A data set is \emph{level-$v$ estimatable} if $v$ is the minimum number such that $\bigcup_{u=0}^v E_u$ equals the set of all the data entries. 
\item If $X$ is level-$v$ estimatable for some $v$, we simply say that $X$ is \emph{estimatable}. If no such $v$ exists, the data set is said to be \emph{unestimatable}.
\end{enumerate}
\end{definition}

We next show the relationship between $E_v$ and $E_{v+1}$, where the proof is available in Appendix B.

\begin{lemma}
\label{lemma:ev_ev1}
If $E_v=\emptyset$, then $E_{v+1}=\emptyset$ for any natural number $v$. 
\end{lemma}

Checking the estimatability of a data set is straightforward because of Lemma~\ref{lemma:ev_ev1}. More specifically,  by Lemma~\ref{lemma:ev_ev1}, a data set $X$ is level-$v$ estimatable if and only if  $v$ is the last index such that $E_v\ne\emptyset$.

In Section \ref{section_model_algorithm}, we present two data imputation algorithms, which rely on different data assumptions. One of the algorithms assumes that the data set is estimatable, while the other algorithm requires a stronger assumption that the data set is level-1 estimatable.

\begin{figure}[ht]
     \begin{center}
        \subfigure[Observed entries ($E_0$)]{%
           \includegraphics[scale=0.8]{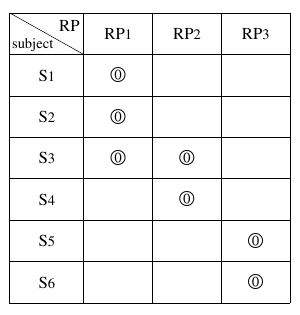} \label{fg_data_properties_example1a_table}
        }\qquad
        \subfigure[$E_0\cup E_1$]{%
          \includegraphics[scale=0.8]{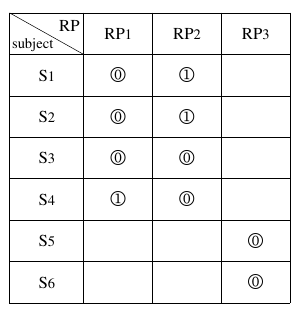} \label{fg_data_properties_example1b_table}
        }
    \end{center}
    \vspace{-0.3cm}
    \caption{Example of unestimatable data}
\label{fg_data_properties_example_table}
\end{figure}

\begin{figure}[ht]
     \begin{center}
        \subfigure[Observed entries ($E_0$)]{%
           \includegraphics[scale=0.8]{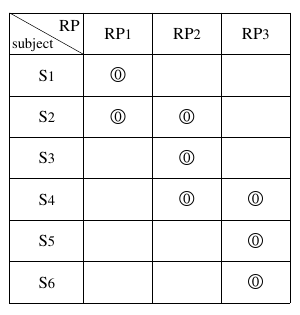} \label{fg_data_properties_example2a_table}
        }\qquad
        \subfigure[$E_0\cup E_1$]{%
          \includegraphics[scale=0.8]{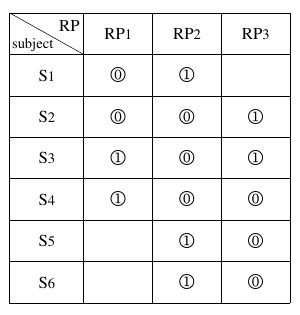} \label{fg_data_properties_example2b_table}
        }\qquad
        \subfigure[$E_0\cup E_1\cup E_2$]{%
          \includegraphics[scale=0.8]{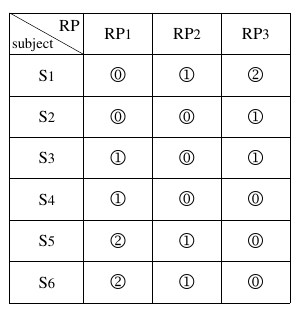} \label{fg_data_properties_example2c_table}
        }
    \end{center}
    
    \vspace{-0.3cm}
    \caption{Example of estimatable data}
\label{fg_data_properties_example_table2}
\end{figure}

\begin{example}
Let us consider the example data sets in Figures \ref{fg_data_properties_example_table} and \ref{fg_data_properties_example_table2}, which present unestimatable and estimatable data sets, respectively. In Figure \ref{fg_data_properties_example_table}, the observed entries are marked with circled 0s. Then, the entry $(i_4,j_1)$ is level-1 estimatable because $(i_3,j_1), (i_3,j_2)$, and $(i_4,j_2)$ are observed. Similarly, the entries $(i_1,j_2)$ and $(i_2,j_2)$ are level-1 estimatable, and we mark them with circled 1s. It can be easily verified that the remaining entries (empty cells in Figure \ref{fg_data_properties_example1b_table}) are unestimatable, indicating that the data set is unestimatable. Notice that the data set does not violate Assumption \ref{assumption_rate_by_one} but violates the estimatability. Let us next consider the data set in Figure~\ref{fg_data_properties_example_table2}. Out of the ten missing entries in Figure \ref{fg_data_properties_example2a_table}, seven are level-1 estimatable. Then, the three remaining entries are level-2 estimatable because they can be estimated based on observed or level-1 estimatable entries. Because all data entries belong to $E_0\cup E_1\cup E_2$, the data set is (level-2) estimatable.\hfill $\square$
\end{example}

We next present a graph characterization of estimatability.
To this end, we introduce several definitions and notations. Consider an undirected graph whose nodes represent RPs and whose edges represent connectivity between nodes.
\begin{definition}
\begin{enumerate}
\item Two RPs are called \emph{connected} if there exists a subject that is rated by both RPs. 
\item Two RPs are called \emph{path-connected} if there exists a path that connects two RPs in the graph.
\item The graph is called \emph{connected} if every pair of RPs is path-connected.
\end{enumerate}
\end{definition}

\begin{figure}[ht]
     \begin{center}
        \subfigure[Graph for Figure~\ref{fg_data_properties_example1a_table}]{%
           \includegraphics[scale=0.5]{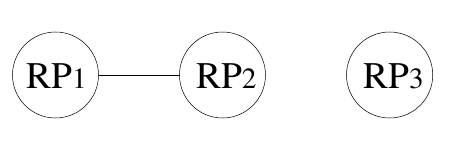} 
        }\qquad \qquad
        \subfigure[Graph for Figure~\ref{fg_data_properties_example2a_table}]{%
          \includegraphics[scale=0.5]{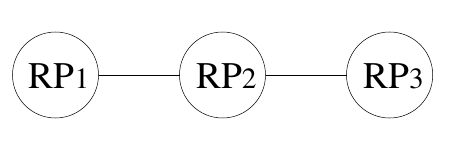} 
        }
    \end{center}
    
    \vspace{-0.3cm}
    \caption{Graph representations for the example data sets}
\label{fg_data_properties_example_graph}
\end{figure}

For an arbitrary set of rating providers $C \subseteq J$, we denote the set of subjects observed by at least one rating provider in $C$ by $\O(C)$. We also denote the set of all the estimatable subjects of the rating providers in  $C$ by $\E(C)$. For a set of single rating provider $\{j\}$, we use shorthand notations $\O(j)$ and $\E(j)$ for $\O(\{j\})$ and $\E(\{j\})$, respectively. The graph representations of data sets in Figures~\ref{fg_data_properties_example1a_table} and \ref{fg_data_properties_example2a_table} are illustrated in Figure~\ref{fg_data_properties_example_graph}. The following results show that the estimatability of a data set can be characterized by the connectivity of the graph representation of the data set.

\begin{theorem}
Given incomplete data $X$ and its graph representation, let $C$ be a connected component and let $\O(C)$ be the set of subjects rated by at least one rating provider in $C$.
Then, all of the entries of the submatrix associated with $\O(C)$ and $C$ are estimatable.
\label{thm_feasibility_check}
\end{theorem}

\begin{theorem}
The input data set is estimatable if and only if its graph representation is connected.
\label{thm_feasibility_check2}
\end{theorem}

The Proofs are available in Appendix B. Estimatability of the input data set will affect the property of the objective function. We discuss this in greater detail in Section~\ref{section_model_algorithm}. By Theorem~\ref{thm_feasibility_check2}, the estimatability check of the input data is equivalent to the connectivity problem for its graph counterpart.
The connectivity can be checked efficiently using various algorithms.
For example, for every distinct pair $(u,v)$ of rating providers, its local connectivity $\kappa(u,v)$ is defined as the maximum number of internally disjoint $u-v$ paths in the graph; it can be efficiently determined using the max-flow min-cut algorithm \citep{dantzig2003max,even1975network}.
Thus, the graph is connected if and only if $\min_{u\ne v}\kappa(u,v)\ne 0$.

One of the algorithms we propose in Section \ref{section_model_algorithm} requires a stronger assumption, level-1 estimatability. We next present a necessary and sufficient condition for an input data set to be level-1 estimatable, where the proof is available in Appendix B.

\begin{theorem}
Given input data $X$, let $I_j$ be the set of subjects that are rated by $j\in J$. For an arbitrary $j'\in J$, let $N_{j'} = \{ j \in J \mid I_j \cap I_{j'} \neq \emptyset \}$ be the set of RPs, including $j'$ itself, which have at least one commonly rated subject with $j'$. Then, $X$ is level-1 estimatable if and only if $I = \bigcup_{j\in N_{j'}} I_j$ for all $j'\in J$. \label{theorem_level1}
\end{theorem}

The key idea of Theorem \ref{theorem_level1} is to check, for each missing entry $(i', j')$, level-1 estimatability can be established if we can find another row and column ($i''$ and $j''$) such that $(i',j''), (i'',j')$, and $(i'', j'')$ are known. The set $N_{j'}$ represents the set of RPs that have a direct connection to $j'$, where the connection is defined by the existence of a commonly rated subject. Next, $\bigcup_{j\in N_{j'}} I_j$ represents the set of subjects rated by at least one neighboring rating provider including $j'$ itself, ensuring the level-1 estimability of the associated missing entries in column $j'$. In other words, $\bigcup_{j \in N_{j'}} I_j$ represents the set of subjects whose entries on column $j'$ are either observed or level-1 estimatable. Therefore, if the union is not equal to the entire subject set $I$, the data set is not level-1 estimatable. If the condition holds for all RPs in $J$, then the data set is level-1 estimatable. We confirm that all the data sets we used in Section \ref{section_experiment} have been tested using Theorem \ref{theorem_level1} and that all are level-1 estimatable.

\subsection{Quadratic Programming Model and Algorithm}

Consider rating matrix $X$ with missing values. To impute its missing values, \cite{kim2019} propose an optimization model that minimizes the number of upsets. This can be written as
\begin{equation}
\min_{\forall x_{kl} \in \Phi} \sum_{(k,l) \in S}  \sum_{(i,j) \in S^{kl}} 1_{(x_{kl} - x_{il})(x_{ij}-x_{kj}) < 0}. \label{def_kim}
\end{equation}
To solve \eqref{def_kim} exactly, the authors reformulate it as a mixed-integer program (MIP) by introducing several auxiliary binary variables. However, the MIP model is not suitable for large-scale data because the problem size increases quadratically in data size: it has $O(m^2n)$ binary variables, $O(m^2n^2)$ continuous variables, and $O(m^2n^2)$ constraints. To overcome computational difficulties and improve scalabilities, the authors use several techniques, including a linear programming relaxation. However, their fastest approach, based on an LP relaxation, is still not sufficiently scalable.

In this paper, we propose a quadratic program (QP) with correlation-based weights to overcome the scalability issue and account for the consensus levels among the RPs.

\begin{equation}
\min_{\forall x_{kl} \in \Phi}  \sum_{(k,l) \in S}  \sum_{(i,j) \in S^{kl}} w_{lj} \Big(\frac{x_{kl}-x_{il}}{c_l} + \frac{x_{ij}-x_{kj}}{c_j} \Big)^2 \label{def_obj}
\end{equation}

Several characteristics of \eqref{def_obj} are notable. First, instead of using the total number of upsets as the objective function, an alternate objective function is used. For row pair $(i,k)$ and column pair $(j,l)$, we adopt the squared value of $\frac{x_{kl}-x_{il}}{c_l} + \frac{x_{ij}-x_{kj}}{c_j}$ instead of the indicator $1_{(x_{kl} - x_{il})(x_{ij}-x_{kj}) < 0}$. The term $(x_{kl}-x_{il})$ captures the difference between the ratings of rows $i$ and $k$ by RP$_l$, while the $(x_{kj}-x_{ij})$ term captures the difference between the ratings of rows $i$ and $k$ by RP$_j$. Then, each term is normalized by dividing by the number of categories $c_l$ and $c_j$. Thus, the entire expression $\frac{x_{kl}-x_{il}}{c_l} + \frac{x_{ij}-x_{kj}}{c_j}$ calculates the rescaled rating discrepancies between RPs $l$ and $j$ for subjects $i$ and $k$. This enables us to avoid a considerable number of binary variables and to keep the problem as an unconstrained continuous optimization, which is easier to solve than MIP. Second, we impose weight $w_{lj}$ between RPs $l$ and $j$. As illustrated in Section \ref{section_analysis}, some RPs show strong concordance, while some RPs create major discordance. We emphasize the ratings from the similar lists for imputing the missing values in the current list by giving higher weights $w_{lj}$ for the comparison between ratings by RPs $j$ and $l$. Note that $w_{lj} \leq 0$ implies zero or negative influences on the objective function, which contradicts our modeling idea. To ensure that the weights positively affect the imputation procedure and to prevent numerical errors due to near zero weights, we truncate weights as $w= \max\{w, \epsilon\}$ where $\epsilon>0$ is a small constant.
Finally, note that the first summation in the objective function considers each missing entry in the matrix, while the second summation considers each entry (regardless of missingness) of the matrix. Hence, each term $\frac{x_{kl}-x_{il}}{c_l} + \frac{x_{ij}-x_{kj}}{c_j}$ has at least one missing entry (i.e., $x_{kl}$), but all four could possibly be missing.

The objective function has a desirable convexity property, as we present in the next theorem with the proof available in Appendix B.

\begin{theorem}
\label{thm:strongconvexity}
Problem \eqref{def_obj} is  convex. In particular, if the input data $X$ includes at least one level-1 estimatable missing entry, then \eqref{def_obj} is strongly convex.
\end{theorem}

Because strong convexity implies strict convexity, we obtain the following corollary. 
\begin{corollary}
If the input data $X$ includes at least one level-1 estimatable missing entry, \eqref{def_obj} has a unique optimal solution. \label{corollary_unique_opt_solution}
\end{corollary}

We note here that, if the input data $X$ is estimatable, then $X$ includes at least one level-1 estimatable missing entry. Therefore, the estimatability of $X$ implies the strong convexity and the existence of a unique solution of \eqref{def_obj}. Because \eqref{def_obj} is strictly convex, multiple solution approaches are available. We can use a general-purpose quadratic programming solver, or we can develop an analytical solution approach based on the first-order optimality condition. Both approaches theoretically guarantee optimality. In the remainder of this section, we develop an analytical solution approach. By setting the partial derivative for all $x_{pq} \in \Phi$ and equating them to zero, we obtain $|Q|$ equations with $|Q|$ unknowns (variables). Because \eqref{def_obj} is convex, the solution to this equation system is the global optimal solution for \eqref{def_obj}. We denote this algorithm by \texttt{QP-AS} and denote the general-purpose solver optimizing \eqref{def_obj} by \texttt{QP-Solver} throughout the remainder of this paper. 

As in the proof of Theorem~\ref{thm:strongconvexity}, without loss of generality, assume that each column of the data is normalized by dividing it by the number of rating categories. Thus, we can conveniently drop the denominators $c_l$'s from the objective function without affecting the analysis. Let $p 
(=|Q|)$ be the number of missing entries. For notational simplicity, we define an index set of missing values $Q^1$ using a one-dimensional index system, whereas $Q$ is based on a two-dimensional index system, so that there is a one-to-one correspondence between $Q$ and $Q^1$. Now, we denote the missing entries by $z_q$ for $q\in Q^1$. We also denote the row and column indices of $z_q$ in the original matrix by $i_q$ and $j_q$. We summarize these new notations as follows.

\begin{enumerate}[noitemsep]
    \item[] $Q^1 = \{1,\cdots,|Q|\}$: index set of missing values
    \item[] $i_q \in I$: row index of $z_q$, $q \in Q^1$, in the original matrix
    \item[] $j_q \in J$: column index of $z_q$, $q \in Q^1$, in the original matrix
\end{enumerate}

For each $q \in Q^1$, where $z_q = x_{i_q j_q}$, we consider all terms in the objective function that include $z_q$. Such a square term corresponds to an entry whose row index is different from $i_q$, while the column index is different from $j_q$. Then, the sum of all such terms is
\[
F = \sum_{j \in J \setminus \{j_q\} }\sum_{i \in I \setminus \{i_q\}} w_{j_q j}(x_{i_qj_q}-x_{i_qj}-x_{ij_q}+x_{ij})^2.
\]
Therefore, the partial derivative of the entire objective function in $z_q$ is equal to $\frac{\partial F}{\partial z_q}$ (or $\frac{\partial F}{\partial x_{i_qj_q}}$), which can be rewritten as follows.
\vspace{0.2cm}

\begin{tabular}{lll}
 $\displaystyle \frac{\partial F}{ \partial z_q}=\frac{\partial F}{ \partial x_{i_qj_q}}$  & = & $\displaystyle 2\sum_{j \in J \setminus \{j_q\} }\sum_{i \in I \setminus \{i_q\}} w_{j_q j}(x_{i_qj_q}-x_{i_qj}-x_{ij_q}+x_{ij})$\\
  & = & $\displaystyle2\sum_{j \in J \setminus \{j_q\}}w_{j_q j}\left[(m-1)x_{i_qj_q}-(m-1)x_{i_qj}-\sum_{i \in I \setminus \{i_q\}} x_{ij_q}+\sum_{i \in I \setminus \{i_q\}} x_{ij}\right]$\\
  & = & $\displaystyle 2 (m-1) \left[\left(\sum_{j \in J \setminus \{j_q\}}w_{j_q j}\right)x_{i_qj_q}-\left(\sum_{j \in J \setminus \{j_q\}}w_{j_qj}x_{i_qj}\right)\right]$\\
  & & $\displaystyle + 2\left[-\left(\sum_{j \in J \setminus \{j_q\}}w_{j_q j}\right)\left(\sum_{i \in I \setminus \{i_q\}}x_{ij_q}\right)+\left(\sum_{i \in I \setminus \{i_q\}}\sum_{j \in J \setminus \{j_q\}} w_{j_qj}x_{ij}\right)\right].$
\end{tabular}
\vspace{0.2cm}

\noindent Then, the equation $\partial F/\partial z_q=0$ can be written as $\sum_{s\in Q^1} A_{qs}x_{i_sj_s}=b_q$ (equivalently, $\sum_{s\in Q^1} A_{qs}z_s=b_q$)
where $A_{qs}$ and $b_q$ are defined as
\begin{equation}
    A_{qs} = \left\{
\begin{array}{cl}
2(m-1)\left(\sum_{j \in J \setminus \{j_q\}}w_{j_q j}\right)&\textup{if $i_q=i_s$ and $j_q=j_s$}\\
-2(m-1)w_{j_q j_s}&\textup{if $i_q=i_s$ and $j_q\ne j_s$}\\
-2\left(\sum_{j \in J \setminus \{j_q\}}w_{j_q j}\right)&\textup{if $i_q\ne i_s$ and $j_q= j_s$}\\
2w_{j_qj_s}&\textup{if $i_q\ne i_s$ and $j_q\ne j_s$}
\end{array}
\right. \label{eqn_A_matrix}
\end{equation}
\begin{equation}
    b_q =      2(m-1)\sum_{j \in J \setminus \{j_q\}} w_{j_q j}y_{i_qj}+2\left(\sum_{j \in J \setminus \{j_q\}}w_{j_q j}\right)\left(\sum_{i \in I \setminus \{i_q\}}y_{ij_q}\right)-2\left(\sum_{i \in I \setminus \{i_q\}}\sum_{j \in J \setminus \{j_q\}} w_{j_qj}y_{ij}\right), \label{eqn_b_vector}
\end{equation}
where $y_{ij}$ for $i\in I, j\in J$ is defined as
\[
y_{ij}=\left\{
\begin{array}{ll}
x_{ij} & \text{if $x_{ij}$ is observed}\\
0 &\text{otherwise}
\end{array}
\right..
\]
The first-order conditions yield a system of equations $Az=b$, where $A$ is a $|Q|\times |Q|$ square matrix. By the strict convexity of the objective function, $A$ is positive definite and hence invertible, implying that the solution to the system is uniquely determined by $z=A^{-1}b$.

Note that $x_{kl} \in \Phi$ in \eqref{def_obj} is a continuous decision variable, while data matrix $X$ may have integer ratings only. When the ratings of $X$ are integers, for both of the proposed algorithms \texttt{QP-AS} and \texttt{QP-solver}, the solution $\hat{X}$ is transformed by $\hat{x}_{kl} = \min \{u_l, \max \{ l_l, round(\hat{x}_{kl}) \} \}$ at the end. This procedure is used for any benchmark algorithm returning continuous values.

Our final remark concerns the computational tricks solving \eqref{def_obj} and preparing \eqref{eqn_A_matrix} and \eqref{eqn_b_vector} more efficiently. Because the ratings are ordinal (and integers), multiple rows (subjects) can be identical, particularly when the number of columns is small. Hence, we can remove the duplicated rows and reformulate \eqref{def_obj} to reduce the problem size while keeping the solutions identical. Let $X'$ be the reduced matrix of $X$ after deleting duplicated rows. $Q'$, $\Phi'$, and $S'$ are defined similarly to $Q$, $\Phi$, and $S$, respectively. Let $d_i$ be the number of duplicated rows identical to row $i$ of $X'$.
\begin{center}
$\displaystyle \min_{\forall x'_{kl} \in \Phi'}  \sum_{(k,l) \in Q'}  \sum_{(i,j) \in S'} w_{lj} d_i d_k \Big(\frac{x_{kl}'-x_{il}'}{c_l} + \frac{x_{ij}'-x_{kj}'}{c_j} \Big)^2$    
\end{center}
Given the normalized matrix $X'$ (obtained by dividing by the number of categories), we can define $A'$ and $b'$ similarly.
\vspace{0.1cm}
\[
    A_{qs}' = \left\{
\begin{array}{cl}
2(\sum_{i \in I \setminus \{i_q\}} d_i d_{i_q})\left(\sum_{j \in J \setminus \{j_q\}}w_{j_q j}\right)&\textup{if $i_q=i_s$ and $j_q=j_s$}\\
-2(\sum_{i \in I \setminus \{i_q\}} d_i d_{i_q})w_{j_q j_s}&\textup{if $i_q=i_s$ and $j_q\ne j_s$}\\
-2 d_{i_q} d_{i_s} \left(\sum_{j \in J \setminus \{j_q\}}w_{j_q j}\right)&\textup{if $i_q\ne i_s$ and $j_q= j_s$}\\
2 d_{i_q} d_{i_s} w_{j_qj_s}&\textup{if $i_q\ne i_s$ and $j_q\ne j_s$}
\end{array}
\right.
\]
\[
 b_q' =      2(\sum_{i \in I \setminus \{i_q\}} d_i d_{i_q})\sum_{j \in J \setminus \{j_q\}} w_{j_q j}y_{i_qj}'+2\left(\sum_{j \in J \setminus \{j_q\}}w_{j_q j}\right)\left(\sum_{i \in I \setminus \{i_q\}} d_i d_{i_q} y_{ij_q}'\right)-2\left(\sum_{i \in I \setminus \{i_q\}}\sum_{j \in J \setminus \{j_q\}} w_{j_qj} d_i d_{i_q} y_{ij}'\right)
\]
\vspace{0.1cm}

We single out the results in this section as a theorem.

\begin{theorem}
The solutions $z=A^{-1}b$ and $z=(A')^{-1}b'$ to the system of equations $Az=b$ and $A'z=b'$, respectively, are the optimal solution to \eqref{def_obj}. \label{theorem_qpas_formula}
\end{theorem}

\subsection{Decomposable Quadratic Programming Model and Algorithm}

Note that \eqref{def_obj} minimizes the total rating discordance using all of the existing and to-be-imputed ratings. In this section, we propose a variant of \eqref{def_obj} to improve the scalability while maintaining a similar solution quality. We simplify the problem by considering the rating discordance for each missing value as follows:

\begin{equation}
\min_{x_{pq} \in \Phi}  \sum_{(p,q) \in Q}  \sum_{(i,j) \in R^{pq}} w_{qj} \Big(\frac{x_{pq}-x_{iq}}{c_q} + \frac{x_{ij}-x_{pj}}{c_j} \Big)^2, \label{def_decomposable_obj}
\end{equation}
where $R^{pq} = \{(i,j) \in S \setminus Q | (i,q) \notin Q, (p,j) \notin Q \}$. Thus, $R_{pq}$ includes $(i,j)$ only if all of $x_{ij}, x_{iq}$, and $x_{pj}$ have available ratings. 
To assure that the optimization problem \eqref{def_decomposable_obj} is well-defined, we assume that the input data $X$ is level-1 estimatable. 

Notice that the level-$1$ estimatability of the input data is equivalent to $R_{pq}\ne\emptyset$ for each $(p,q)\in Q$. The objective function of \eqref{def_decomposable_obj} is also strongly convex due to the existence of a level-1 estimatable entry. We omit its proof because it is almost identical to that of Theorem~\ref{thm:strongconvexity}.
Therefore, the solution resulting from the first-order condition is the unique global optimal solution.

Notice that the objective function of \eqref{def_decomposable_obj} includes no bilinear term.
Therefore, each partial derivative in $x_{pq}$ does not include any other decision variables, implying that the $x_{pq}$ component of the optimal solution can be obtained directly by finding the zero of its partial derivative in $x_{pq}$.
More precisely, for fixed $p$ and $q$, \eqref{def_decomposable_obj} includes only one decision variable, i.e., $x_{pq}$ and existing ratings. Therefore, \eqref{def_decomposable_obj} can be decomposed into $|Q|$ sub-problems, where each sub-problem tries to impute one missing value in $\Phi$. In detail, for each $x_{pq} \in \Phi$, we define the following optimization problem.

\begin{equation}
\min_{x_{pq}}   \sum_{(i,j) \in R_{pq}} w_{qj} \Big(\frac{x_{pq}-x_{iq}}{c_q} + \frac{x_{ij}-x_{pj}}{c_j} \Big)^2 \label{def_obj_svi}
\end{equation}

We can solve \eqref{def_obj_svi} based on the same derivative-based approach calculating the imputed value from the following equation.
\begin{equation*}
\sum_{(i,j) \in R_{pq}} w_{qj} \Big(\frac{x_{pq}-x_{iq}}{c_q} + \frac{x_{ij}-x_{pj}}{c_j} \Big) = 0
\end{equation*}
The closed form formula is
\begin{equation}
\displaystyle x_{pq}^* =  \frac{\sum_{(i,j) \in R_{pq}} w_{qj} \Big(\frac{x_{iq}}{c_q} + \frac{x_{pj}-x_{ij}}{c_j} \Big)}{\sum_{(i,j) \in R_{pq}} \big(\frac{w_{qj}}{c_q} \big)}. \label{def_decomposable_obj_formula}
\end{equation}
\begin{theorem} 
The solution $x^*$ in \eqref{def_decomposable_obj_formula} is the optimal solution to \eqref{def_decomposable_obj}. \label{theorem_dqpsvas_formula}
\end{theorem}
We denote \eqref{def_decomposable_obj} by \texttt{dQP} and denote the single value analytical solution approach solves \eqref{def_obj_svi} for each $x_{pq}$ by \texttt{dQP-SVAS}. Finally, 
Table \ref{tab_algo_summary} summarizes all proposed models and algorithms in this section. To deal with the drawbacks of the MIP model of \cite{kim2019}, we do not impose integrality constraints, but instead approximately minimize upsets in the imputed data by solving QP \eqref{def_obj}, where positive Kendall rank correlation matrices in Figure \ref{fg_kendall} led us to weight the upsets. The QP \eqref{def_obj} can be solved by a general-purpose solver (\texttt{QP-Solver}) or analytical solution approach (\texttt{QP-AS}). By imputing each missing value iteratively (\texttt{dQP-SVAS}), the solution time is improved.

\begin{table}[htbp]
  \centering
 \begin{small}
        \begin{tabular}{|c|c|c|c|l|}
    \hline
        Algorithm & Model reference & Solver or formula & Sufficient condition  & \multicolumn{1}{c|}{Relevant theorems} \\ \hline
    QP-Solver & \eqref{def_obj}     & General purpose solver  & Level-1 estimatability & Theorem \ref{thm:strongconvexity}, Corollary \ref{corollary_unique_opt_solution} \\ 
    QP-AS & \eqref{def_obj}     & \eqref{eqn_A_matrix}, \eqref{eqn_b_vector}   &    Level-1 estimatability   & Theorem \ref{theorem_qpas_formula} \\ 
    dQP-SVAS & \eqref{def_decomposable_obj}, \eqref{def_obj_svi}  & \eqref{def_decomposable_obj_formula}     &   Level-1 estimatability    & Theorem \ref{theorem_dqpsvas_formula} \\ \hline
    \end{tabular}%
 \end{small}
  \caption{Summary of Proposed Models and Algorithms}
  \label{tab_algo_summary}%
\end{table}%

\section{Computational Experiment}
\label{section_experiment}

In this section, we compare the performance of the proposed algorithms (\texttt{QP-solver}, \texttt{QP-AS}, and \texttt{dQP-SVAS}) with multiple benchmarks: \texttt{cart}, \texttt{LP} relaxation of \cite{kim2019}, and \texttt{softImpute}. We select the three benchmarks after testing various methods in the literature whose R packages are available, including VIM.hotdeck, ClustImpute.ClustImpute, ECLRMC.ECLRMC, mice.polr, mice.pmm, and mic.sample, where the package and function names are separated by the dot. \texttt{LP} and \texttt{QP-solver} are implemented in C\# and solved by CPLEX 20.1. \texttt{QP-AS} and \texttt{dQP-SVAS} are implemented and tested with R 4.0.4. The codes for \texttt{QP-AS} and \texttt{dQP-SVAS} are available in the online supplement. For the benchmarks, \texttt{cart} from the R package \textit{mice} \citep{buuren2011mice} and \texttt{softImpute} from the R package \textit{softImpute} \citep{hastie2015softimpute, hastie2015matrix, mazumder2010spectral} are tested with R 4.0.4. We adopted $\epsilon=0.01$ as the truncating bound for the weights $w_{lj}$.
For all experiments, a personal computer with 32 GB RAM and Intel Core i7-10700 CPU @ 2.90 GHz was used.

\subsection{Experimental Data}
In this section, we describe the data sets used in the experiment. Starting from the input data (with some missing values for real data and without missing values for synthetic data), we delete some of the known ratings so that we can check the imputation quality based on the deleted ratings. All real data sets used in the experiment are available in the online supplement.

\subsubsection{Real Data}
We generate six families of experimental data sets based on the six original real-world data sets summarized in Table \ref{table_experiment_data}. All of the test data sets were generated based on the procedure used by \cite{kim2019}. We briefly describe the procedure as follows. For each of the data sets listed in Table \ref{table_experiment_data}, after randomly partitioning the known ratings into ten groups, we create each instance by deleting the known ratings in exactly one group. Hence, each 10-fold instance has additional missing values compared to the raw matrix in Table \ref{table_experiment_data}, while the true values of the deleted missing cells are known so that we can assess the model performance.

\begin{table}[htbp]

  \centering
  \begin{small}
    \begin{tabular}{|c|c|c|c|c|}
    \hline
    Data set  & Shorthand & Size  & Number of  & Master \\
    name  & name  & $(m,n)$      & instances & matrix \\ \hline
    Hospital 10-fold & Hospital10F & (4217,4) & 10    & Hospital in Table \ref{table_data_summary} \\
    Journal  10-fold& Journal10F  & (914,13) & 10    & Journal in Table \ref{table_data_summary} \\
    ESG 10-fold & ESG10F & (1356,4) & 10 & ESG in Table \ref{table_data_summary}\\
    Elementary School 10-fold & Elementary10F & (301,5) & 10    & Elementary School in Table \ref{table_data_summary} \\
    High School 10-fold & Highschool10F & (132,5) & 10    & High School in Table \ref{table_data_summary} \\
    Movielens 10-fold & Movielens10F & (1102,12) & 10 & Movielens in Table \ref{table_data_summary} \\
    \hline
    \end{tabular}%
\end{small}    
    \caption{Experimental data sets}
  \label{table_experiment_data}%
\end{table}%

\subsubsection{Synthetic Data}

We generate synthetic data using Algorithm \ref{algo_synthetic_data}. In Step 1, a random correlation matrix is generated in which all correlations are between $s-0.2$ and $s+0.2$, while $s$ is the input correlation parameter. In Steps 2 and 3, a multivariate normal data set is created using the correlation matrix in Step 1. The generated continuous values are converted into $1-5$ scale ratings using the procedure described in Section \ref{section_data_prep_conversion}. The generated rating matrix $X$ at the end of Step 3 is saved as the true matrix. Note that $X_{true}$ in Step 4 does not have missing values. Hence, in Steps 5-11, we selectively delete rating values to create the final synthetic data $X$ with missing ratings. In Step 6, given column $j$, we first define selection weights $\Omega$ for each entry, where poorly rated entries have smaller weights to be selected. For example, if $x_{1j} = 5$ (good) and $x_{2j}=1$ (bad), the weights to be sampled are 1 and 5, respectively, which indicates the latter has a higher probability to be deleted in the next step. In Step 7, \textit{round(rm)} entries are randomly selected and deleted based on the selection weights $\Omega$. This procedure in Steps 6 and 7 is based on the observation in Section \ref{subsec_analysis_missing_value_patterns}. Finally, in Steps 8-10, we make sure that each row has at least one rating available, satisfying Assumption \ref{assumption_rate_by_one}.

 \begin{algorithm}[h!] 
     \caption{Synthetic data generation}
     \begin{algorithmic}[1] \label{algo_synthetic_data}
     \REQUIRE $m,n,s,r$ (number of rows, number of columns, input correlation, missing rates)
     \STATE $\Sigma$ $\gets$ Generate a random $n \times n$ correlation matrix with correlations between $s-0.2$ and $s+0.2$
     \STATE $X$ $\gets$ Generate an $m \times n$ multivariate normal data with zero means and correlation matrix $\Sigma$
     \STATE $X$ $\gets$ Convert the continuous values of $X$ into ratings with five categories
     \STATE $X_{true} \gets X$
     \STATE \textbf{for} each column $j \in J$
     \STATE \quad Define $\Omega = \{\omega_i \in \mathbb{R}^+| i \in I\}$, where $\omega_i = 6-x_{ij}$ represents the weight of entry $i$ for random selection
     \STATE \quad Randomly select and delete $round(rm)$ entries of the column $j\in J$ based on selection weights $\Omega$
     \STATE \textbf{end for}
     \STATE \textbf{for} each row $i \in I$
     \STATE \quad \textbf{if} all values in row $i$ are missing \textbf{then} randomly select a column from row $i$ of $X_{true}$ and add it to $X$
     \STATE \textbf{end for}
   \end{algorithmic}
 \end{algorithm}

We generate 10 instances for each tuple $(m,n,s,r)$, where $m \in \{1000,2000, 3000\}$, $n \in \{6,8,10\}$, $s \in \{0.3,0.5,0.7\}$, and $r \in \{0.2,0.3,0.4\}$. Hence, there are 810 $(=3^4 \times 10)$ synthetic instances, which are used for the experiment in Section~\ref{subsection_experiment_result}. We make a few remarks regarding the generated data sets. First, the choice of $s$, which generates positive correlation matrices, is supported by the analysis in Section~\ref{subsec_analysis_consensus_level}, where we show that all RPs are positively correlated in all real data sets. Furthermore, the set of $s$ covers a reasonable range for correlations. Second, the correlation parameter $s$ mostly leads to lower Kendall rank correlations in the final data, as we convert the continuous values into integers and delete ratings. The final Kendall rank correlation matrices show less positively correlated columns. Finally, because the missing rates of all RPs are strictly less than 0.5, any pair of RPs has at least one subject rated by both of the PRs, and the corresponding nodes in the graph representation are connected. Therefore, by Theorem \ref{thm_feasibility_check2}, the generated data set is estimatable. In Appendix C, we present an analysis on the probability that the data set generated by Algorithm \ref{algo_synthetic_data} is estimatable for a more general missing rate greater than 0.5. The analysis shows that Algorithm \ref{algo_synthetic_data} generates estimatable data sets with a high probability for reasonable values of $m,n,$ and $r$. This implies that a few repetitions of Algorithm \ref{algo_synthetic_data} can return an estimatable data set successfully if estimatability is required.

\subsection{Performance Measures and Result}
\label{subsection_experiment_result}
Recall that, for any original data set $X$, $Q$ is the index set of the missing entries of $X$. Because we delete some of the known ratings in $X$ to generate the test data sets, there are more missing values in the test data sets. For a 10-fold test data set, let $Q^*$ be the index set of missing ratings that are deleted and let $r_{kl}$ be the deleted true known ratings for $x_{kl}$, $(k,l) \in Q^*$. Then, for each 10-fold data set, we have missing values in $Q^* \cup Q$, whereas $Q^* \cap Q = \emptyset$. In other words, because we delete the values in $Q^*$ from the original data set $X$, each 10-fold data set will have missing values that are deleted (from $Q^*$), as well as missing values that were originally missing (from $Q$).

To measure the performance, we use the following four metrics.
\begin{enumerate}[noitemsep]
    \item Time, running time in seconds.
    \item Accuracy, defined as $ \big( \sum_{(k,l) \in Q^*} 1_{x_{kl} = r_{kl}} \big) / |Q^*|$, measures the proportion of the correctly imputed values among the missing values with known ratings.
    \item Root Mean Square Error (RMSE), defined as $ \sqrt{\big( \sum_{(k,l) \in Q^*} (x_{kl} - r_{kl})^2 \big) / |Q^*|}$, measures the square root of the average squared errors of the imputed values among the missing values with known ratings.
    \item Mean Absolute Deviation (MAD), defined as $ \big( \sum_{(k,l) \in Q^*} |x_{kl} - r_{kl}| \big) / |Q^*|$, measures the average absolute errors of the imputed values among the missing values with known ratings.
\end{enumerate}

For the real data sets, we additionally check how the Kendall rank correlation matrices are changed before and after the imputation. Let $T^B$ and $\hat{T}^B$ be the Kendall rank correlation matrices of original and imputed data matrices for all RP pairs, where $T^B_{j_1 j_2}$ and $\hat{T}^B_{j_1 j_2}$ represent the element for RP $j_1$ and $j_2$. Recall that Figure \ref{fg_kendall} presents $T^B$. Because all elements of the $T_B$ and $\hat{T}_B$ have very small p-values for all algorithms, we use the following three metrics to measure the deviations.
\begin{enumerate}[noitemsep]
    \item RMSE$_{\tau}$: $\sqrt{\sum_{(j_1,j_2) \in J} (\hat{T}^B_{j_1 j_2} -T_{j_1 j_2})^2/ \binom{n}{2}}$
    \item MAD$_{\tau}$: $\sum_{(j_1,j_2) \in J} |\hat{T}^B_{j_1 j_2} -T_{j_1 j_2}| / \binom{n}{2}$
    \item AvgD$_{\tau}$: $\sum_{(j_1,j_2) \in J} (\hat{T}^B_{j_1 j_2} -T_{j_1 j_2}) / \binom{n}{2}$
\end{enumerate}
Note that, by comparing MAD$_{\tau}$ and AvgD$_{\tau}$, we can observe if the imputed matrices have mostly positive or negative changes.

\subsubsection{Result for Real Data}

In Table \ref{tab:result_real}, the result for the real data sets is presented. For each data set, the average performances out of 10 instances are reported. For Accuracy, RMSE, and MAD, the gray cells indicate that the corresponding algorithm's output is the best out of all algorithms compared; the boldface fonts indicate that the relative gap of the corresponding algorithm's output is within 5\% from the best value. For the Journal10F data set, \texttt{LP} did not solve the problems due to the memory issue, and \texttt{QP-solver} used over 20GB of memory on average. For the Movielens10F data set, both \texttt{LP} and \texttt{QP-solver} had the same memory issue.

\begin{table}[htbp]
  \centering
  \begin{scriptsize}
    \begin{tabular}{|c|c|c|c|c|c|c|c|c|}
    \hline
    Data set & Measure & cart  & softImpute    & missForest & LP    & QP-solver & QP-AS & dQP-SVAS \\ \hline
    Hospital10F & Time  & 3.7   & 0.0   & 3.1   & 54.9  & 4.8   & 0.8   & 0.8 \\
          & Accuracy & 0.3460 & 0.3498 & 0.4054 & \cellcolor[rgb]{ .851,  .851,  .851}\textbf{0.4333} & \textbf{0.4240} & \textbf{0.4240} & \textbf{0.4248} \\
          & RMSE  & 1.1969 & 1.3556 & \textbf{0.9588} & \textbf{0.9780} & \cellcolor[rgb]{ .851,  .851,  .851}\textbf{0.9559} & \cellcolor[rgb]{ .851,  .851,  .851}\textbf{0.9559} & \textbf{0.9560} \\
          & MAD   & 0.8891 & 0.9695 & \textbf{0.6996} & \textbf{0.6896} & \textbf{0.6841} & \textbf{0.6841} & \cellcolor[rgb]{ .851,  .851,  .851}\textbf{0.6833} \\ \hline
    Journal10F & Time  & 3.8   & 0.1   & 5.1   & NA    & 615.8 & 13.8  & 1.8 \\
          & Accuracy & 0.6967 & 0.6950 & \cellcolor[rgb]{ .851,  .851,  .851}\textbf{0.7663} & NA    & 0.6801 & 0.6801 & 0.6898 \\
          & RMSE  & 0.6218 & 0.6021 & \cellcolor[rgb]{ .851,  .851,  .851}\textbf{0.5131} & NA    & 0.5970 & 0.5970 & 0.5886 \\
          & MAD   & 0.3298 & 0.3235 & \cellcolor[rgb]{ .851,  .851,  .851}\textbf{0.2438} & NA    & 0.3318 & 0.3318 & 0.3220 \\ \hline
    ESG10F & Time  & 1.1   & 0.0   & 0.9   & 10.3  & 1.5   & 0.1   & 0.1 \\
          & Accuracy & 0.4600 & \cellcolor[rgb]{ .851,  .851,  .851}\textbf{0.4858} & 0.4480 & \textbf{0.4773} & \textbf{0.4711} & \textbf{0.4711} & 0.4564 \\
          & RMSE  & 1.1895 & 1.1977 & 1.0918 & 1.1725 & \cellcolor[rgb]{ .851,  .851,  .851}\textbf{0.9637} & \cellcolor[rgb]{ .851,  .851,  .851}\textbf{0.9637} & 1.0337 \\
          & MAD   & 0.7840 & 0.7649 & 0.7316 & 0.7609 & \cellcolor[rgb]{ .851,  .851,  .851}\textbf{0.6498} & \cellcolor[rgb]{ .851,  .851,  .851}\textbf{0.6498} & 0.6978 \\ \hline
    Elementary10F & Time  & 0.6   & 0.0   & 0.3   & 4.9   & 1.3   & 0.1   & 0.0 \\
          & Accuracy & 0.5386 & 0.5193 & \textbf{0.6145} & \textbf{0.6349} & \cellcolor[rgb]{ .851,  .851,  .851}\textbf{0.6386} & \textbf{0.6373} & \textbf{0.6361} \\
          & RMSE  & 0.8336 & 1.0382 & 0.6996 & \textbf{0.6610} & \textbf{0.6443} & \textbf{0.6453} & \cellcolor[rgb]{ .851,  .851,  .851}\textbf{0.6441} \\
          & MAD   & 0.5337 & 0.6398 & 0.4193 & \textbf{0.3892} & \cellcolor[rgb]{ .851,  .851,  .851}\textbf{0.3807} & \textbf{0.3819} & \textbf{0.3819} \\ \hline
    Highschool10F & Time  & 0.5   & 0.0   & 0.1   & 0.3   & 0.2   & 0.0   & 0.0 \\
          & Accuracy & 0.5682 & 0.4045 & \textbf{0.6818} & \cellcolor[rgb]{ .851,  .851,  .851}\textbf{0.6818} & \textbf{0.6591} & \textbf{0.6591} & \textbf{0.6500} \\
          & RMSE  & 0.8799 & 1.0658 & \textbf{0.6749} & \textbf{0.6546} & \cellcolor[rgb]{ .851,  .851,  .851}\textbf{0.6467} & \cellcolor[rgb]{ .851,  .851,  .851}\textbf{0.6467} & \textbf{0.6743} \\
          & MAD   & 0.5409 & 0.7500 & \textbf{0.3682} & \cellcolor[rgb]{ .851,  .851,  .851}\textbf{0.3591} & \textbf{0.3682} & \textbf{0.3682} & 0.3864 \\ \hline
    Movielens10F & Time  & 5.0   & 0.1   & 5.5   & NA    & NA    & 46.4  & 2.4 \\
          & Accuracy & 0.3565 & 0.4202 & \textbf{0.4486} & NA    & NA    & \cellcolor[rgb]{ .851,  .851,  .851}\textbf{0.4678} & \textbf{0.4668} \\
          & RMSE  & 1.2033 & 1.0654 & 0.9431 & NA    & NA    & \textbf{0.8993} & \cellcolor[rgb]{ .851,  .851,  .851}\textbf{0.8968} \\
          & MAD   & 0.8782 & 0.7418 & 0.6570 & NA    & NA    & \textbf{0.6204} & \cellcolor[rgb]{ .851,  .851,  .851}\textbf{0.6196} \\ \hline
    \end{tabular}%
        \end{scriptsize}
  \caption{Real data result (boldface = within 5\% relative gap, graycell = best)} \label{tab:result_real}%
\end{table}%

From Table \ref{tab:result_real}, we observe that the imputation qualities (measured by accuracy, RMSE, and MAD) of \texttt{missForest}, \texttt{LP}, \texttt{QP-solver}, \texttt{QP-AS}, and \texttt{dQP-SVAS} are in the best group, while \texttt{cart} and \texttt{softImpute} are in the second group. Generally, the proposed methods (\texttt{QP-solver}, \texttt{QP-AS}, and \texttt{dQP-SVAS}) provide the best or near-best solutions, except for Journal10F. The \texttt{missForest} method performs exceptionally well for the Journal10F data set, although it occasionally fails to provide a best or near-best solution for the other data sets. In terms of running time, \texttt{softImpute} is the fastest method, followed by \texttt{sQP-SVAS} and \texttt{cart}. However, \texttt{softImpute} has the worst overall performance regarding other performance measures. For most data sets, \texttt{LP} performs well in terms of imputation qualities (accuracy, RMSE, and MAD), although the computation time is poor. Overall, the running time differences among the algorithms are negligible (except for \texttt{LP}) for the real data sets.

\begin{figure}[ht]
     \begin{center}
        \subfigure[Hospital10F]{%
           \includegraphics[width=0.27\textwidth]{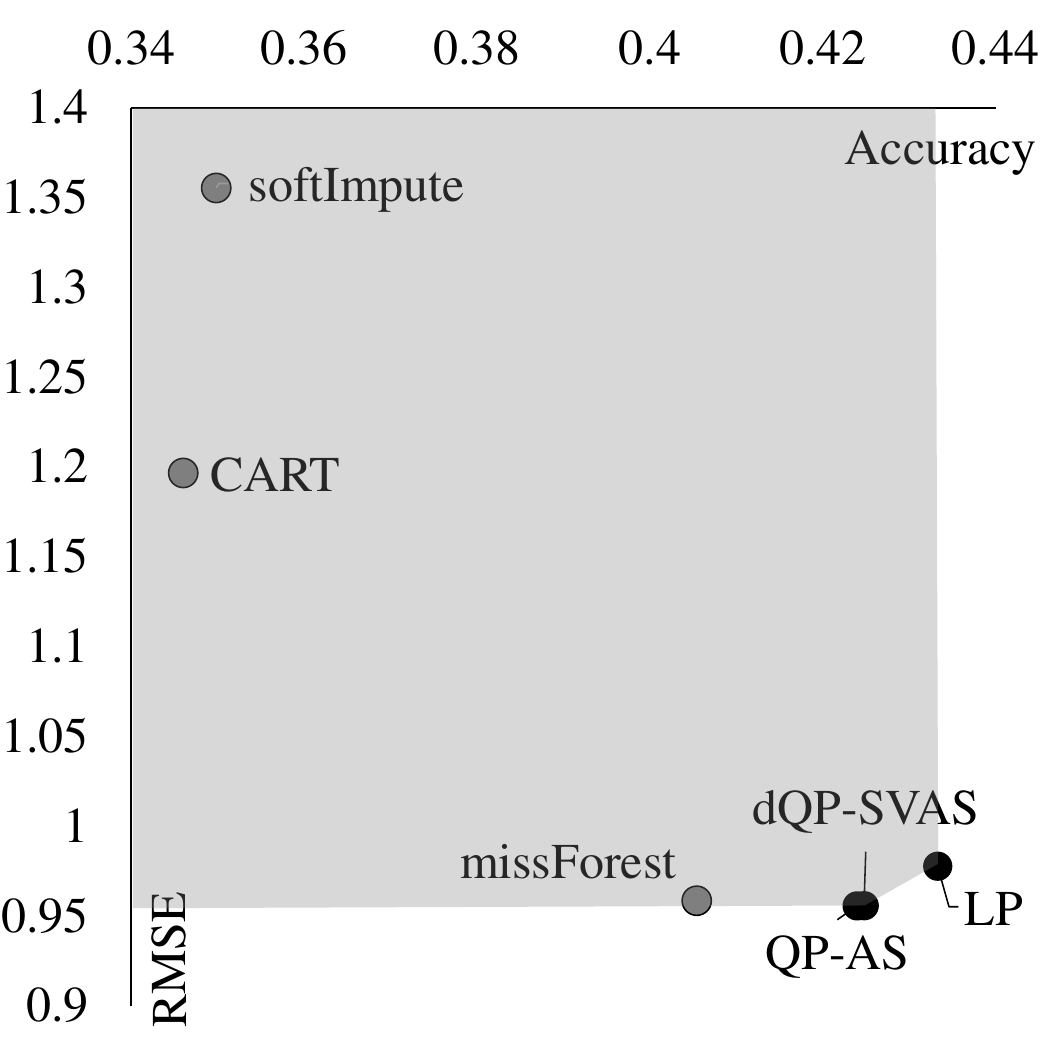} \label{hospitalscatt}
        }\quad
        \subfigure[Journal10F]{%
          \includegraphics[width=0.27\textwidth]{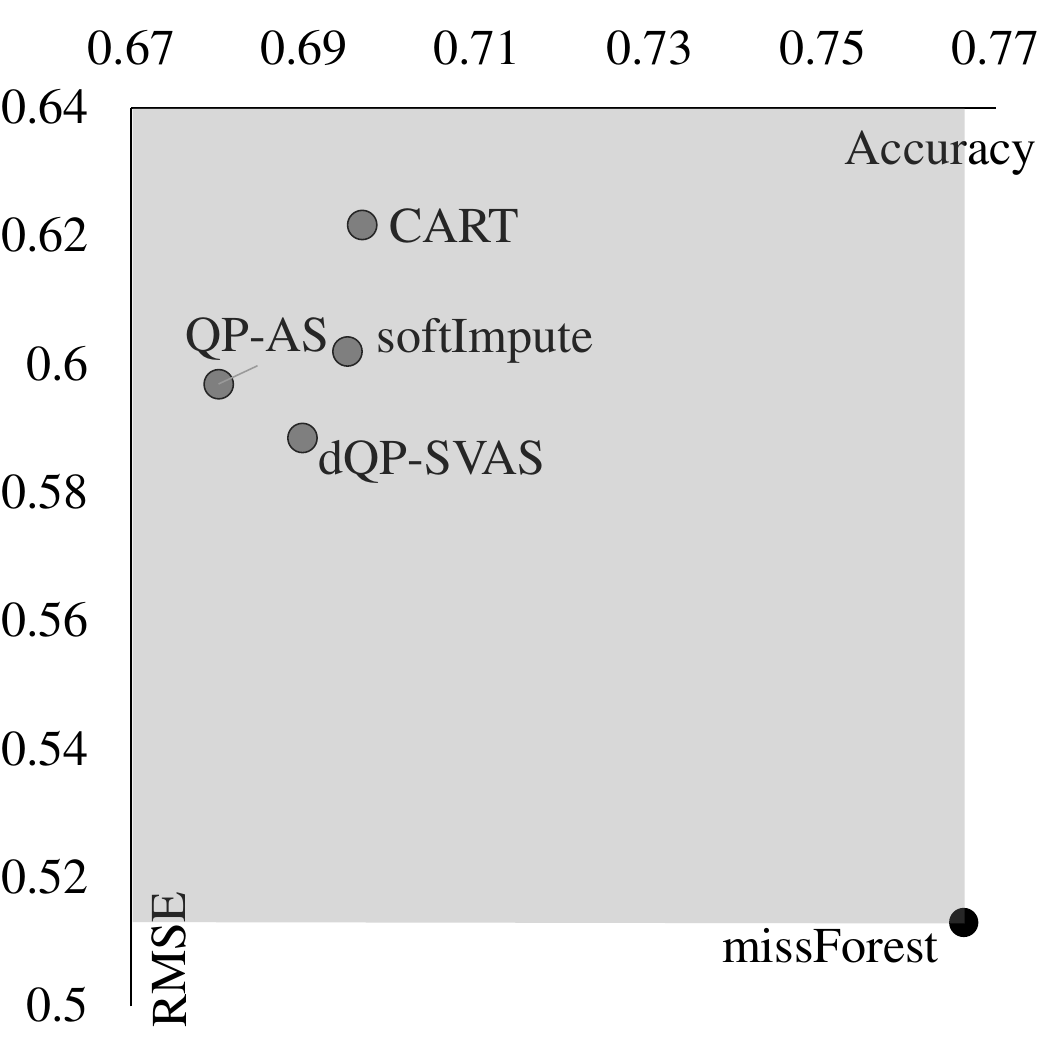} \label{journalscatt}
        }\quad
        \subfigure[ESG10F]{%
          \includegraphics[width=0.27\textwidth]{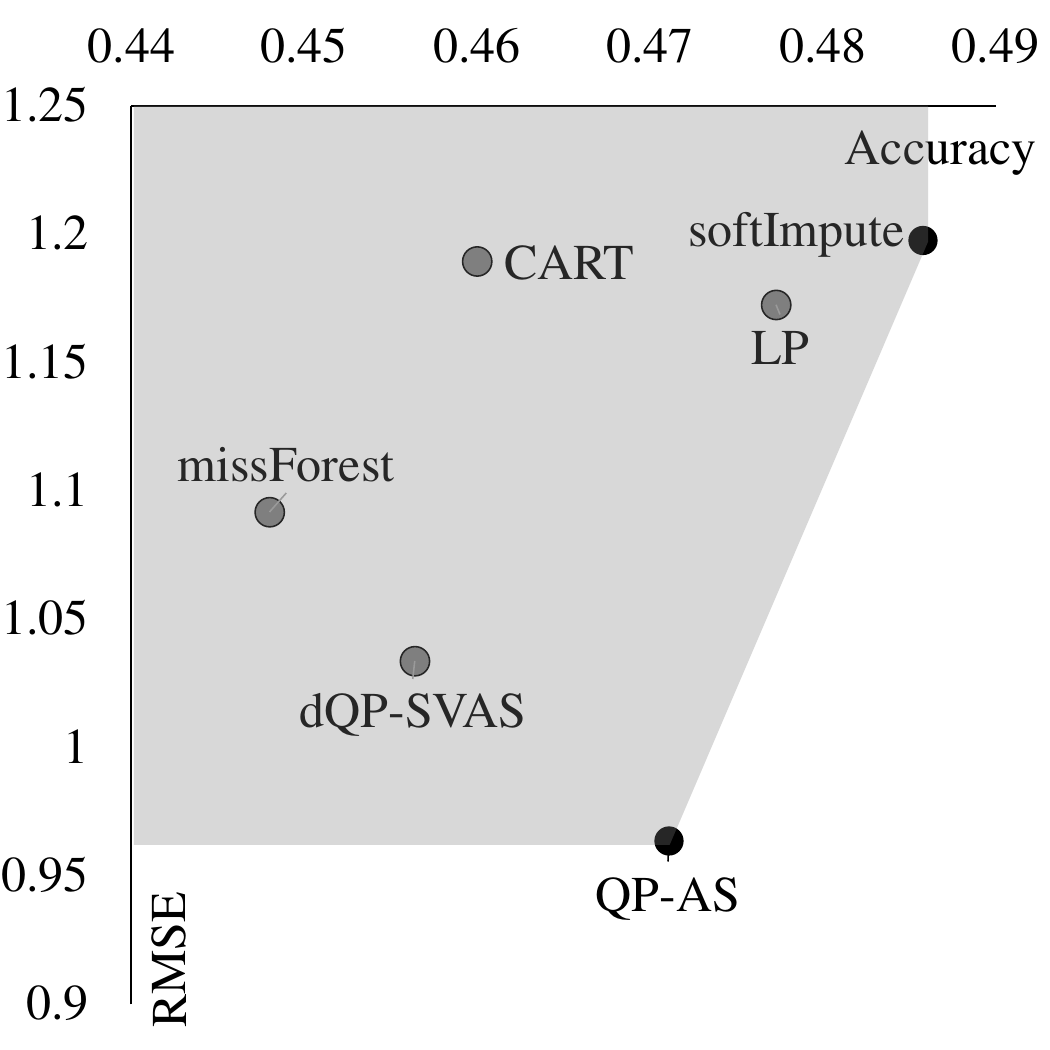} \label{esgscatt}
        }\\
        \subfigure[Elementary10F]{%
           \includegraphics[width=0.27\textwidth]{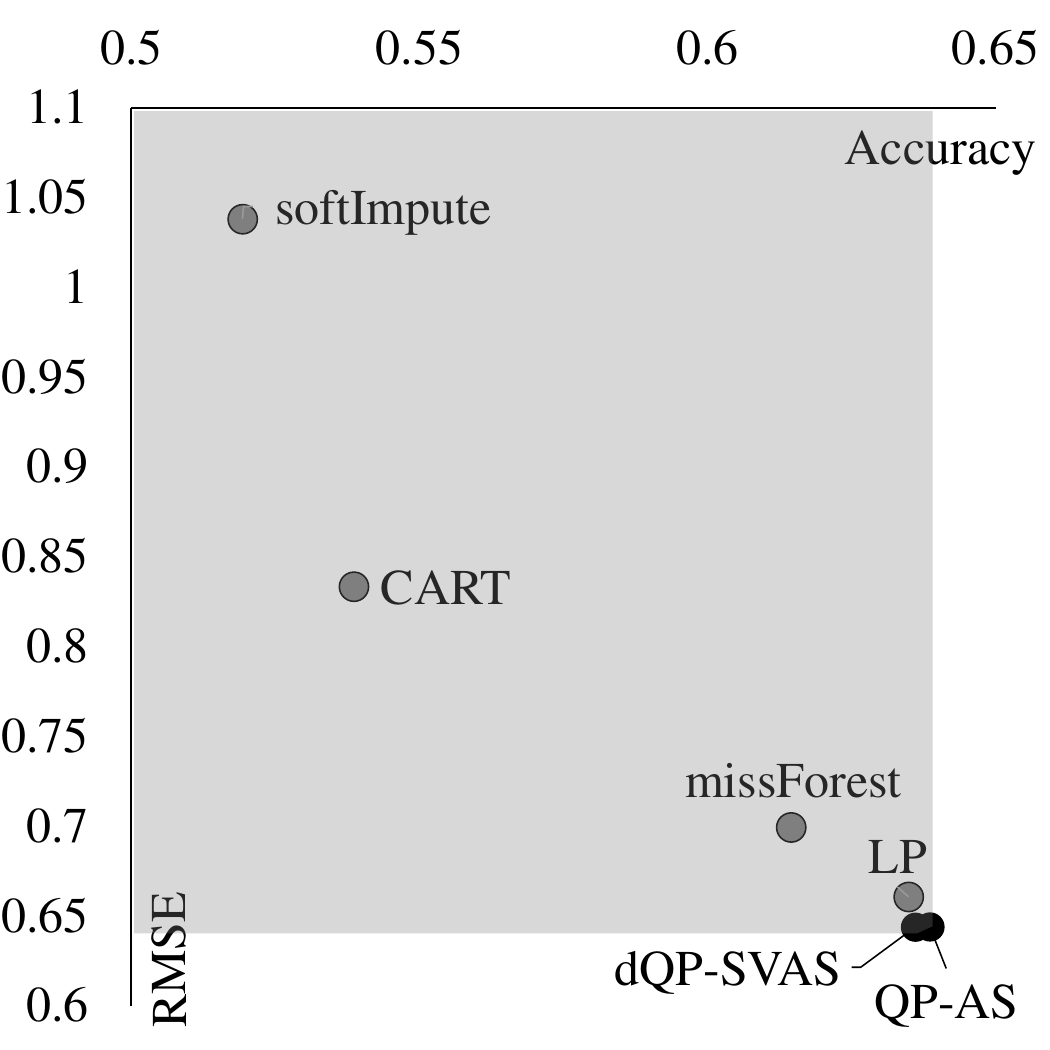} \label{elementaryscatt}
        }\quad
        \subfigure[Highschool10F]{%
          \includegraphics[width=0.27\textwidth]{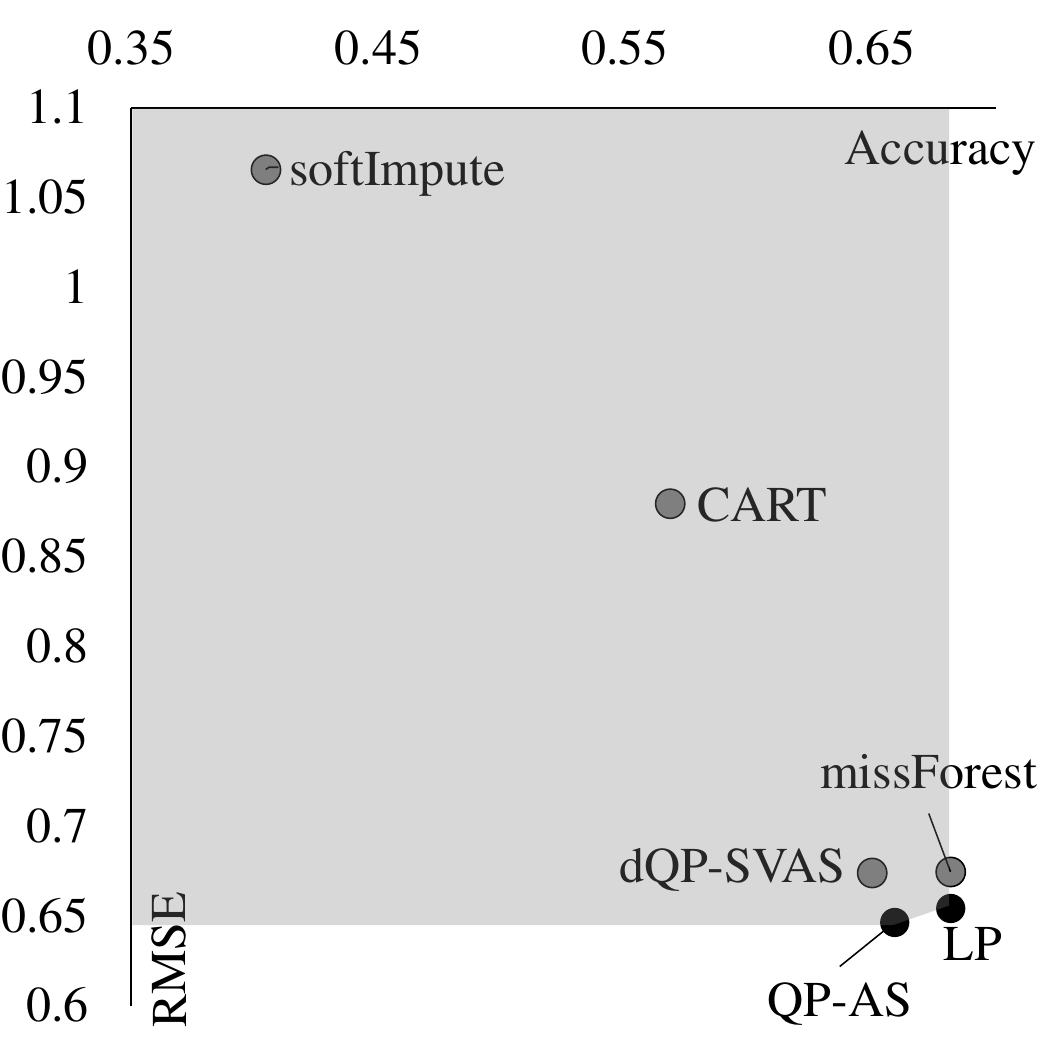} \label{highschoolscatt}
        }\quad
        \subfigure[Synthetic]{%
          \includegraphics[width=0.27\textwidth]{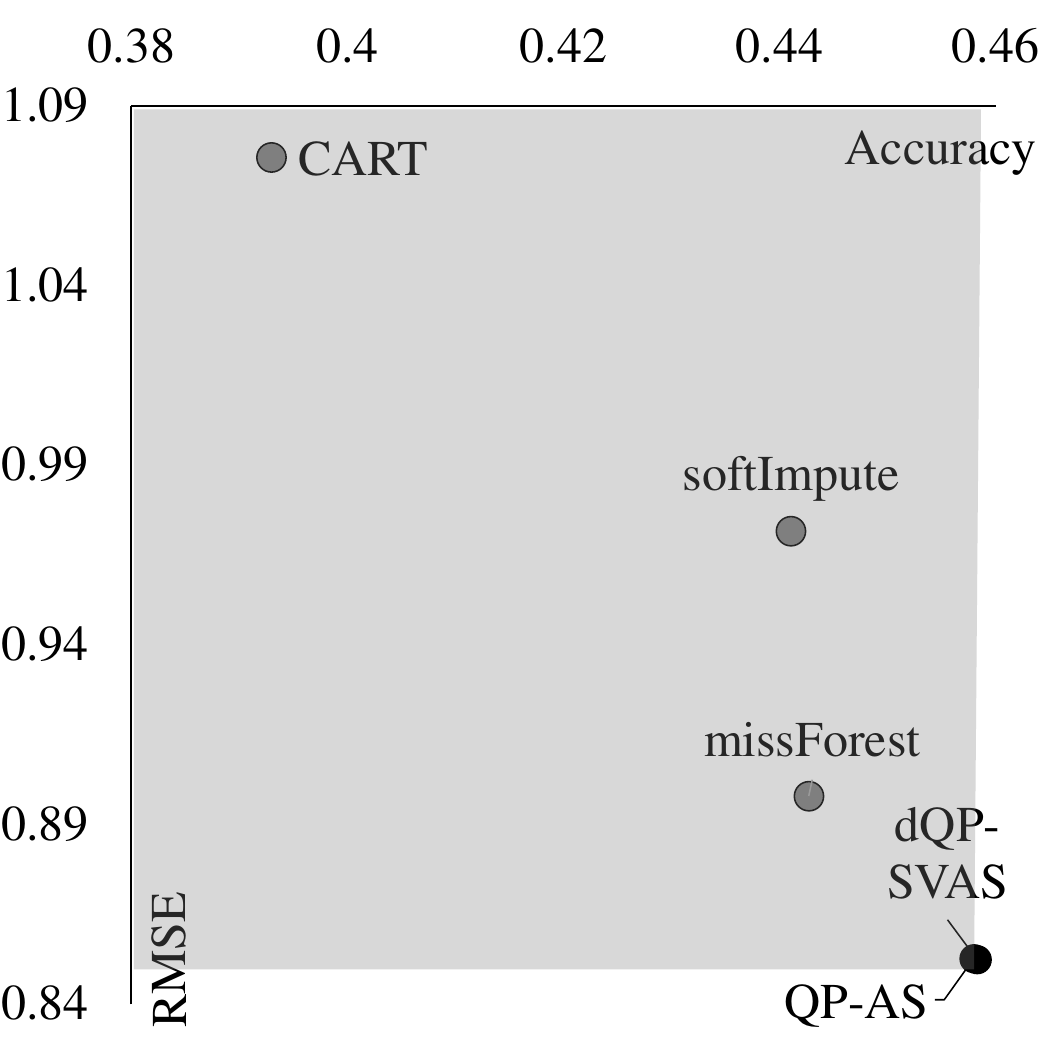} \label{syntheticscatt}
        }           
    \end{center}
    
    \vspace{-0.3cm}
    \caption{Relative comparison with accuracy and RMSE}
\label{fg_pareto_opt}
\end{figure}
We next focus on comparing the solution quality. In Figure \ref{fg_pareto_opt}, we present scatter plots of accuracy and RMSE for each family of the test data sets. \texttt{QP-solver} is excluded here because \texttt{QP-AS} provides almost identical results with a much faster running time. In each plot, the horizontal and vertical axes represent the accuracy and RMSE. We indicate the Pareto optimal algorithms, which are not dominated by the other algorithms, with black colored circles. The areas dominated by the Pareto optimal points are indicated by gray color (excludes the boundary); we indicate the algorithms in the dominated area using gray-colored circles. The result shown in Figure \ref{fg_pareto_opt} indicates that the proposed algorithms \texttt{QP-AS} and \texttt{dQP-SVAS} are on or near the Pareto optimal boundaries, except for the Journal10 data set. \texttt{missForest} is also on or near the Pareto optimal boundaries, except for the ESG10F data set.
\texttt{cart} and \texttt{softImpute} are frequently far away from the boundaries. Overall, we conclude that \texttt{QP-AS}, \texttt{dQP-SVAS}, and \texttt{missForest} outperform \texttt{cart} and \texttt{softImpute} in the real-world data sets.

In Table \ref{tab:result_real_kendall}, the result of Kendall rank correlation matrix comparison is presented. Note that the original six data set in Table \ref{table_data_summary} are used instead of the 10-fold data sets in Table \ref{table_experiment_data}. Hence, each algorithm solves each data set exactly once. The second column presents the number of RP pairs, calculated by $\binom{n}{2}$. For all performance measures, the gray cells indicate that the corresponding algorithm's output is the best out of all algorithms compared, where the minimum is the best for RMSE$_\tau$ and MAD$_\tau$ and the close-to-zero is the best for AvgD$_\tau$. The boldface fonts indicate that the relative gap of the corresponding algorithm's output is within 5\% from the best value. The result indicates that \texttt{cart} is the best in general, which is different from the result in Table \ref{tab:result_real}. \texttt{missForest} performs best or near best for the Journal and Highschool data sets, and \texttt{QP-AS} and \texttt{dQP-SVAS} outperforms for the Journal data set. Comparisons of MAD$_\tau$ and AvgD$_\tau$ brings an interesting observation. Because \texttt{QP-AS} and \texttt{dQP-SVAS} minimize upsets, which is indirectly maximizing the Kendall rank correlation, MAD$_\tau$ and AvgD$_\tau$ are almost identical for \texttt{QP-AS} and \texttt{dQP-SVAS}. In other words, the Kendall rank correlations increase after the imputation by \texttt{QP-AS} and \texttt{dQP-SVAS}. All other benchmark algorithms that are not based on upset minimization have mixed signs for AvgD$_\tau$.

\begin{table}[htbp]
  \centering
  \begin{scriptsize}
        \begin{tabular}{|c|c|c|c|c|c|c|c|c|c|}
        \hline
   Data set       & \#pairs & measure & cart  & softImpute & missForest & LP    & QP-solver & QP-AS & dQP-SVAS \\ \hline
    Hospital & 6     & RMSE$_\tau$  & \cellcolor[rgb]{ .749,  .749,  .749}0.0119 & 0.3713 & 0.1100 & 0.2652 & 0.2425 & 0.2425 & 0.2427 \\
          &       & MAD$_\tau$   & \cellcolor[rgb]{ .749,  .749,  .749}0.0085 & 0.3444 & 0.0878 & 0.2490 & 0.2275 & 0.2275 & 0.2277 \\
          &       & AvgD$_\tau$  & \cellcolor[rgb]{ .749,  .749,  .749}0.0072 & -0.1099 & 0.0878 & 0.2490 & 0.2275 & 0.2275 & 0.2277 \\ \hline
    Journal & 55    & RMSE$_\tau$  & 0.1436 & 0.1857 & 0.0872 & NA    & \textbf{0.0854} & \textbf{0.0854} & \cellcolor[rgb]{ .749,  .749,  .749}0.0829 \\
          &       & MAD$_\tau$   & 0.1218 & 0.1304 & \cellcolor[rgb]{ .749,  .749,  .749}0.0690 & NA    & \textbf{0.0709} & \textbf{0.0709} & \textbf{0.0694} \\
          &       & AvgD$_\tau$  & -0.1211 & -0.1001 & -0.0558 & NA    & \textbf{0.0254} & \textbf{0.0254} & \cellcolor[rgb]{ .749,  .749,  .749}0.0236 \\ \hline
    ESG   & 6     & RMSE$_\tau$  & \cellcolor[rgb]{ .749,  .749,  .749}0.0706 & 0.0928 & 0.0889 & 0.2704 & 0.2248 & 0.2248 & 0.2364 \\
          &       & MAD$_\tau$   & \cellcolor[rgb]{ .749,  .749,  .749}0.0533 & 0.0820 & 0.0731 & 0.2476 & 0.2002 & 0.2002 & 0.2194 \\
          &       & AvgD$_\tau$  & -0.0421 & 0.0677 & \cellcolor[rgb]{ .749,  .749,  .749}-0.0377 & 0.2476 & 0.2002 & 0.2002 & 0.2194 \\ \hline
    Elementary & 10    & RMSE$_\tau$  & \cellcolor[rgb]{ .749,  .749,  .749}0.0562 & 0.3867 & 0.0941 & 0.1337 & 0.1374 & 0.1374 & 0.1361 \\
          &       & MAD$_\tau$   & \cellcolor[rgb]{ .749,  .749,  .749}0.0402 & 0.3350 & 0.0847 & 0.1109 & 0.1179 & 0.1179 & 0.1161 \\
          &       & AvgD$_\tau$  & \cellcolor[rgb]{ .749,  .749,  .749}-0.0390 & -0.2479 & 0.0809 & 0.1069 & 0.1179 & 0.1179 & 0.1161 \\ \hline
    Highschool & 10    & RMSE$_\tau$  & 0.4344 & 0.5396 & \cellcolor[rgb]{ .749,  .749,  .749}0.0666 & 0.1284 & 0.1218 & 0.1233 & 0.1200 \\
          &       & MAD$_\tau$   & 0.3773 & 0.4235 & \cellcolor[rgb]{ .749,  .749,  .749}0.0584 & 0.1090 & 0.1066 & 0.1080 & 0.1037 \\
          &       & AvgD$_\tau$  & -0.3773 & -0.3905 & 0.0156 & \cellcolor[rgb]{ .749,  .749,  .749}-0.0002 & 0.0827 & 0.0844 & 0.0847 \\ \hline
    Movielens & 66    & RMSE$_\tau$  & \cellcolor[rgb]{ .749,  .749,  .749}0.0486 & 0.2755 & 0.2279 & NA    & NA    & 0.5332 & 0.5229 \\
          &       & MAD$_\tau$   & \cellcolor[rgb]{ .749,  .749,  .749}0.0387 & 0.2388 & 0.2159 & NA    & NA    & 0.5256 & 0.5150 \\
          &       & AvgD$_\tau$  & \cellcolor[rgb]{ .749,  .749,  .749}-0.0198 & 0.1861 & 0.2159 & NA    & NA    & 0.5256 & 0.5150 \\ \hline
    \end{tabular}%
  \end{scriptsize}
  \caption{Real data Kendall rank correlation result (boldface = within 5\% relative gap, graycell = best)} \label{tab:result_real_kendall}%
\end{table}%

Considering all results in this section, we conclude that prediction quality is not closely related to the Kendall rank correlation matrix similarity.  \texttt{QP-AS}, \texttt{dQP-SVAS}, and \texttt{missForest} outperform in terms of prediction, whereas \texttt{cart} outperforms in keeping the Kendall rank correlation matrix similar. When the missing values occur randomly, and the Kendall rank correlation matrix is believed to be true, we recommend using \texttt{cart} to retain a similar correlation matrix after the imputation. However, when the missing values do not occur randomly, and there is doubt in the Kendall rank correlation matrix, we recommend using our proposed algorithms for obtaining more accurate imputed values.

We finally remark that the proposed algorithms, \texttt{QP-AS} and \texttt{dQP-SVAS}, are deterministic. Hence, given a fixed data set, they always return the same set of imputed values and do not address the uncertainty issues of missing value imputation. To check the robustness of the proposed algorithms over varying subsets of the samples, we present multiple imputation versions of the proposed algorithms in Appendix D. Multiple imputation generates multiple plausible values for each missing value to create multiple complete data sets \citep{rubin1996multiple}. With iterative random samplings of the rows, the multiple imputation versions of the proposed algorithms can return multiple imputed values for each missing value. The results in Appendix D indicate \texttt{QP-AS} and \texttt{dQP-SVAS} return consistent imputed values. Also, the prediction performances are slightly better than the corresponding multiple imputation versions, while the solution times of \texttt{QP-AS} and \texttt{dQP-SVAS} are much faster.

\subsubsection{Result for Synthetic Data}

In this section, we present the results for the synthetic data sets in Tables \ref{tab:synthetic_m} - \ref{tab:synthetic_missrate}. Note that \texttt{LP} and \texttt{QP-solver} are excluded in the comparison due to the scalability issue. Furthermore, because there are 810 instances in total, the aggregated results by $m$ (number of subjects), $n$ (number of RPs), correlation parameter $s$, and missing rate are presented in Tables \ref{tab:synthetic_m}, \ref{tab:synthetic_n}, \ref{tab:synthetic_corr}, and \ref{tab:synthetic_missrate}, respectively. Similar to the format in Table \ref{tab:result_real}, the best algorithm and the algorithms within 5\% of the best are indicated by gray cells and boldfaces, respectively. In all tables, we observe that \texttt{dQP-SVAS} outperforms the other methods in most cases. The running time of \texttt{dQP-SVAS} is the second-best after \texttt{softImpute}, while the solution quality of \texttt{dQP-SVAS} is decisively the best. The solution quality of \texttt{QP-AS} is frequently within 5\% of the best, while \texttt{softImpute} and \texttt{missForest} occasionally provide good solutions. Overall, \texttt{dQP-SVAS} and \texttt{QP-AS} provide lower RMSEs and MADs.

 \begin{table}[htbp]
  \centering
  \begin{scriptsize}
    \begin{tabular}{|c|c|c|c|c|c|c|}
    \hline
    $m$     & Measure & cart  & softImpute    & missForest & QP-AS & dQP-SVAS \\ \hline
    1000  & Time  & 5.68  & 0.04  & 19.04 & 6.16  & 1.15 \\
          & Accuracy & 0.3934 & \textbf{0.4405} & \textbf{0.4424} & \textbf{0.4575} & \cellcolor[rgb]{ .851,  .851,  .851}\textbf{0.4577} \\
          & RMSE  & 1.0739 & 0.9762 & 0.8985 & \textbf{0.8540} & \cellcolor[rgb]{ .851,  .851,  .851}\textbf{0.8538} \\
          & MAD   & 0.7797 & 0.6839 & 0.6397 & \textbf{0.6047} & \cellcolor[rgb]{ .851,  .851,  .851}\textbf{0.6044} \\ \hline
    2000  & Time  & 5.91  & 0.04  & 19.91 & 32.94 & 3.98 \\
          & Accuracy & 0.3931 & \textbf{0.4414} & \textbf{0.4431} & \textbf{0.4586} & \cellcolor[rgb]{ .851,  .851,  .851}\textbf{0.4589} \\
          & RMSE  & 1.0755 & 0.9707 & 0.8973 & \textbf{0.8514} & \cellcolor[rgb]{ .851,  .851,  .851}\textbf{0.8511} \\
          & MAD   & 0.7808 & 0.6800 & 0.6385 & \textbf{0.6025} & \cellcolor[rgb]{ .851,  .851,  .851}\textbf{0.6021} \\ \hline
    3000  & Time  & 6.18  & 0.04  & 21.24 & 93.96 & 8.55 \\
          & Accuracy & 0.3924 & \textbf{0.4413} & \textbf{0.4427} & \textbf{0.4579} & \cellcolor[rgb]{ .851,  .851,  .851}\textbf{0.4583} \\
          & RMSE  & 1.0777 & 0.9681 & 0.8979 & \textbf{0.8527} & \cellcolor[rgb]{ .851,  .851,  .851}\textbf{0.8524} \\
          & MAD   & 0.7827 & 0.6787 & 0.6390 & \textbf{0.6037} & \cellcolor[rgb]{ .851,  .851,  .851}\textbf{0.6032} \\ \hline
    \end{tabular}%
\end{scriptsize}
  \caption{Synthetic data results (averaged by $m$, number of subjects)}
  \label{tab:synthetic_m}%
\end{table}%

Table \ref{tab:synthetic_m} shows that, when the results are aggregated by $m$, \texttt{softImpute}, \texttt{missForest}, \texttt{QP-AS}, and \texttt{dQP-SVAS} provide similar accuracy. The solution quality changes very little as $m$ changes for all algorithms. The solution time of \texttt{softImpute} is constantly the best, followed by \texttt{dQP-SVAS}. However, as $m$ increases, the running time of \texttt{dQP-SVAS} increases more rapidly than \texttt{cart} and \texttt{missForest}. Hence, we expect that the running time of \texttt{dQP-SVAS} will be slower than that of \texttt{cart} and \texttt{missForest} when $m$ is large.

\begin{table}[htbp]
  \centering
  \begin{scriptsize}
    \begin{tabular}{|c|c|c|c|c|c|c|}
    \hline
    $n$     & Measure & cart  & softImpute    & missForest & QP-AS & dQP-SVAS \\ \hline
    6     & Time  & 5.90  & 0.04  & 19.87 & 7.26  & 1.60 \\
          & Accuracy & 0.3932 & 0.4414 & 0.4428 & \textbf{0.4682} & \cellcolor[rgb]{ .851,  .851,  .851}\textbf{0.4684} \\
          & RMSE  & 1.0754 & 0.9707 & 0.8978 & \textbf{0.8415} & \cellcolor[rgb]{ .851,  .851,  .851}\textbf{0.8413} \\
          & MAD   & 0.7807 & 0.6801 & 0.6389 & \textbf{0.5910} & \cellcolor[rgb]{ .851,  .851,  .851}\textbf{0.5907} \\ \hline
    8     & Time  & 6.25  & 0.04  & 21.63 & 38.01 & 4.43 \\
          & Accuracy & 0.3932 & \textbf{0.4410} & \textbf{0.4426} & \textbf{0.4578} & \cellcolor[rgb]{ .851,  .851,  .851}\textbf{0.4581} \\
          & RMSE  & 1.0754 & 0.9720 & 0.8983 & \textbf{0.8517} & \cellcolor[rgb]{ .851,  .851,  .851}\textbf{0.8517} \\
          & MAD   & 0.7807 & 0.6811 & 0.6394 & \textbf{0.6032} & \cellcolor[rgb]{ .851,  .851,  .851}\textbf{0.6030} \\ \hline
    10    & Time  & 5.61  & 0.04  & 18.68 & 87.80 & 7.66 \\
          & Accuracy & 0.3925 & \textbf{0.4408} & \textbf{0.4428} & \textbf{0.4480} & \cellcolor[rgb]{ .851,  .851,  .851}\textbf{0.4484} \\
          & RMSE  & 1.0764 & 0.9724 & \textbf{0.8976} & \textbf{0.8649} & \cellcolor[rgb]{ .851,  .851,  .851}\textbf{0.8643} \\
          & MAD   & 0.7818 & 0.6814 & \textbf{0.6389} & \textbf{0.6167} & \cellcolor[rgb]{ .851,  .851,  .851}\textbf{0.6161} \\ \hline
    \end{tabular}%
\end{scriptsize}
  \caption{Synthetic data results (averaged by $n$, number of RPs)}
  \label{tab:synthetic_n}%
\end{table}%

In Table \ref{tab:synthetic_n}, when the results are aggregated by $n$, \texttt{QP-AS} and \texttt{dQP-SVAS} consistently perform well, while \texttt{softImpute} and \texttt{missForest} provide competitive accuracy when $n = 8$ and 10. We observe that the accuracies of \texttt{QP-AS} and \texttt{dQP-SVAS} decrease as $n$ increases, while the benchmark algorithms' accuracies remain relatively constant as $n$ changes. In terms of the running time, we observe the same trend. \texttt{softImpute} is the fastest, while the running time of \texttt{dQP-SVAS} increases much quickly than \texttt{cart} and \texttt{missForest} as $n$ increases.

The result in Table \ref{tab:synthetic_corr} shows that the solution quality of \texttt{QP-AS} and \texttt{dQP-SVAS} improves drastically as the correlation parameter $s$ increases. The other methods provide relatively constant solution quality in increasing $s$. Therefore, we conclude that the proposed algorithms perform well when the ratings are highly correlated.

\begin{table}[htbp]
  \centering
  \begin{scriptsize}
    \begin{tabular}{|c|c|c|c|c|c|c|}
    \hline
    corr     & Measure & cart  & softImpute    & missForest & QP-AS & dQP-SVAS \\ \hline
    0.3   & Time  & 5.93  & 0.04  & 20.09 & 44.80 & 4.62 \\
          & Accuracy & 0.3905 & \textbf{0.4367} & \cellcolor[rgb]{ .851,  .851,  .851}\textbf{0.4393} & \textbf{0.4186} & \textbf{0.4188} \\
          & RMSE  & 1.0784 & 0.9770 & \textbf{0.9009} & \textbf{0.8843} & \cellcolor[rgb]{ .851,  .851,  .851}\textbf{0.8842} \\
          & MAD   & 0.7845 & 0.6871 & \cellcolor[rgb]{ .851,  .851,  .851}\textbf{0.6431} & \textbf{0.6478} & \textbf{0.6475} \\ \hline
    0.5   & Time  & 5.92  & 0.04  & 20.10 & 44.89 & 4.60 \\
          & Accuracy & 0.3920 & \textbf{0.4402} & \textbf{0.4418} & \textbf{0.4505} & \cellcolor[rgb]{ .851,  .851,  .851}\textbf{0.4509} \\
          & RMSE  & 1.0776 & 0.9732 & \textbf{0.8994} & \textbf{0.8634} & \cellcolor[rgb]{ .851,  .851,  .851}\textbf{0.8631} \\
          & MAD   & 0.7829 & 0.6823 & \textbf{0.6405} & \textbf{0.6142} & \cellcolor[rgb]{ .851,  .851,  .851}\textbf{0.6138} \\ \hline
    0.7   & Time  & 5.90  & 0.04  & 19.99 & 43.37 & 4.46 \\
          & Accuracy & 0.3964 & 0.4463 & 0.4471 & \textbf{0.5049} & \cellcolor[rgb]{ .851,  .851,  .851}\textbf{0.5051} \\
          & RMSE  & 1.0710 & 0.9648 & 0.8935 & \textbf{0.8104} & \cellcolor[rgb]{ .851,  .851,  .851}\textbf{0.8100} \\
          & MAD   & 0.7756 & 0.6732 & 0.6335 & \textbf{0.5488} & \cellcolor[rgb]{ .851,  .851,  .851}\textbf{0.5484} \\ \hline
    \end{tabular}%
\end{scriptsize}
  \caption{Synthetic data results (averaged by correlation parameter $s$)}
  \label{tab:synthetic_corr}%
\end{table}%

The result in Table \ref{tab:synthetic_missrate} shows that the solution quality of \texttt{QP-AS} and \texttt{dQP-SVAS} decreases as the missing rate increases, while other methods provide relatively constant solution quality. However, the effect of the missing rate on the solution quality is less significant than the effect of $s$.

\begin{table}[htbp]
  \centering
  \begin{scriptsize}
    \begin{tabular}{|c|c|c|c|c|c|c|}
    \hline
    missrate     & Measure & cart  & softImpute    & missForest & QP-AS & dQP-SVAS \\ \hline
    0.2   & Time  & 5.94  & 0.04  & 20.23 & 15.61 & 3.94 \\
          & Accuracy & 0.3943 & \textbf{0.4424} & \textbf{0.4437} & \textbf{0.4623} & \cellcolor[rgb]{ .851,  .851,  .851}\textbf{0.4623} \\
          & RMSE  & 1.0731 & 0.9671 & 0.8950 & \cellcolor[rgb]{ .851,  .851,  .851}\textbf{0.8366} & \textbf{0.8368} \\
          & MAD   & 0.7785 & 0.6774 & 0.6368 & \cellcolor[rgb]{ .851,  .851,  .851}\textbf{0.5918} & \textbf{0.5919} \\ \hline
    0.3   & Time  & 5.92  & 0.04  & 20.09 & 39.85 & 4.66 \\
          & Accuracy & 0.3930 & \textbf{0.4411} & \textbf{0.4427} & \textbf{0.4582} & \cellcolor[rgb]{ .851,  .851,  .851}\textbf{0.4588} \\
          & RMSE  & 1.0755 & 0.9718 & 0.8979 & \textbf{0.8511} & \cellcolor[rgb]{ .851,  .851,  .851}\textbf{0.8507} \\
          & MAD   & 0.7809 & 0.6809 & 0.6391 & \textbf{0.6025} & \cellcolor[rgb]{ .851,  .851,  .851}\textbf{0.6019} \\ \hline
    0.4   & Time  & 5.90  & 0.04  & 19.87 & 77.61 & 5.09 \\
          & Accuracy & 0.3916 & \textbf{0.4397} & \textbf{0.4417} & \textbf{0.4535} & \cellcolor[rgb]{ .851,  .851,  .851}\textbf{0.4538} \\
          & RMSE  & 1.0785 & 0.9761 & \textbf{0.9008} & \textbf{0.8705} & \cellcolor[rgb]{ .851,  .851,  .851}\textbf{0.8699} \\
          & MAD   & 0.7837 & 0.6844 & \textbf{0.6414} & \textbf{0.6166} & \cellcolor[rgb]{ .851,  .851,  .851}\textbf{0.6160} \\ \hline
    \end{tabular}%
\end{scriptsize}
  \caption{Synthetic data results (averaged by missing rate)}
  \label{tab:synthetic_missrate}%
\end{table}%

\begin{figure}[ht]
     \begin{center}
        \subfigure[$n$ (num RPs)]{%
           \includegraphics[scale=0.5]{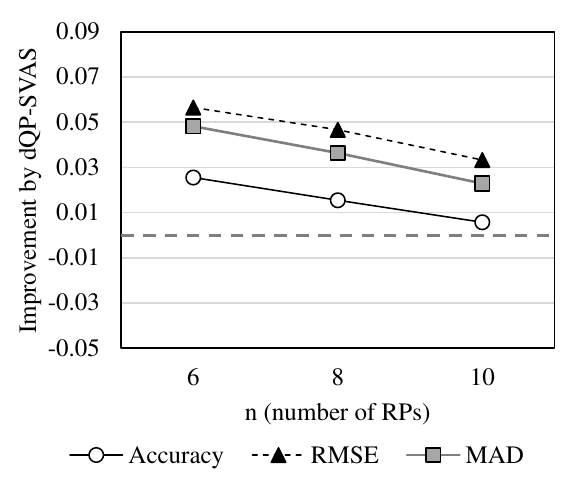} \label{fg_svas_perf_n}
        }\quad
        \subfigure[correlation]{%
          \includegraphics[scale=0.5]{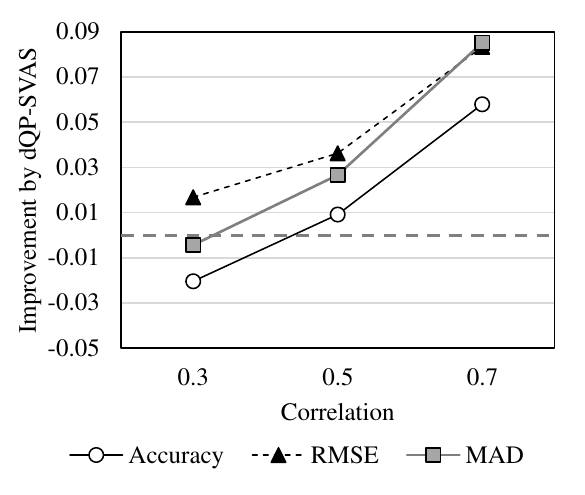} \label{fg_svas_perf_corr}
        }\quad
        \subfigure[missing rate]{%
          \includegraphics[scale=0.5]{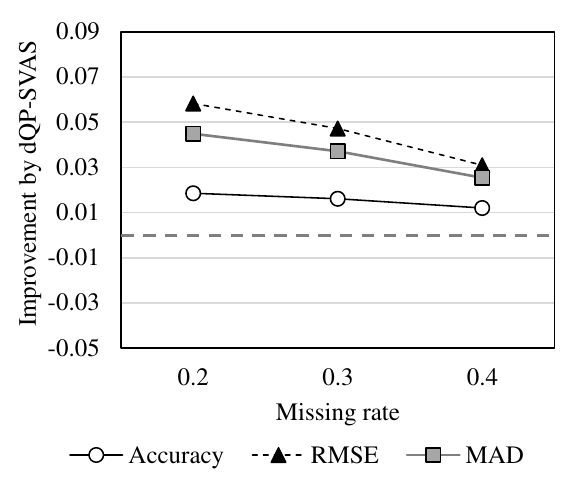} \label{fg_svas_perf_missrate}
        }
    \end{center}
    
    \vspace{-0.5cm}
    \caption{Improvement by dQP-SVAS}
\label{fg_svas_perf}
\end{figure}

To investigate the effect of $n$, $s$, and the missing rate in greater detail, we compare in Figure \ref{fg_svas_perf} the performance of \texttt{dQP-SVAS} against the three benchmark algorithms in the literature (\texttt{cart}, \texttt{softImpute}, and \texttt{missForest}). In each plot, the horizontal axis represents a data characteristic such as $n$, correlation, and missing rates, while the vertical axis represents the improvement by \texttt{dQP-SVAS} from the best outcome out of the three benchmarks, which is defined as
\begin{itemize}
    \item[] obj$_{\texttt{dQP-SVAS}}$ - max$\{$obj$_{\texttt{cart}}$,obj$_{\texttt{softImpute}}$,obj$_{\texttt{missForest}}\}$, for obj = accuracy,
    \item[] min$\{$obj$_{\texttt{cart}}$,obj$_{\texttt{softImpute}}$,obj$_{\texttt{missForest}}\}$ - obj$_{\texttt{dQP-SVAS}}$, for obj = RMSE, MAD.
\end{itemize}
If the improvement is positive, this implies that \texttt{dQP-SVAS} outperforms. Otherwise, it implies that one of the three algorithms defeats \texttt{dQP-SVAS}.

The plots in Figure \ref{fg_svas_perf} summarize the relative performance of \texttt{dQP-SVAS} in increasing $n$, correlation, and the missing rate. As $n$ increases, the solution quality improvement by \texttt{dQP-SVAS} from the three algorithms decreases. Thus, we expect a worse performance for \texttt{dQP-SVAS} when $n$ is large, which explains the poor performance of \texttt{dQP-SVAS} (and other proposed algorithms) for the Journal10F data set in Table \ref{tab:result_real}. As the missing rate increases, \texttt{dQP-SVAS} becomes less attractive, although it outperforms the three benchmarks in all performance measures for missing rates 0.2, 0.3, and 0.4. The effect of correlation parameter $s$ is shown in Figure \ref{fg_svas_perf_corr}. As correlations among RPs increase, the improvement by \texttt{dQP-SVAS} from the three algorithms rapidly increases. Therefore, we conclude that \texttt{dQP-SVAS} performs particularly well for the input data with a few highly correlated PRs and a relatively small number of missing values.

\section{Conclusion}

Ratings provide succinct and credible measures for comparing subjects and are used throughout individual and corporate decision-making processes. However, missing entries are frequently observed in a combined list. Hence, accurate estimation or imputation of the missing entries in the combined rating data sets can address this problem and significantly impact users' decision-making processes.

In this paper, we propose three algorithms and examine the properties of missing values in the combined rating list. In detail, we present quadratic programming models and algorithms tailored for missing rating imputation. The proposed algorithms outperform the state-of-the-art general imputation methods such as \texttt{cart}, \texttt{missForest}, and \texttt{softImpute} in terms of imputation accuracy, while showing outstanding scalability compared with the MIP-based rating imputation model reported in the literature. The proposed algorithms tend to increase the Kendall rank correlation coefficients after the imputation, but we observe that preserving the original data's Kendall rank correlation coefficients does not necessarily guarantee imputation quality. The proposed algorithms can also be used for imputing continuous values, with a slight modification on normalization. Our algorithms perform particularly well when the columns are highly correlated.
We also provide a procedure and definitions for defining when the imputation algorithms are applicable. These analyses show that the consensus level between the RPs varies across contexts, that MAR assumptions are frequently violated, and that missingness may indicate superiority or inferiority of the subject. The synthetic data sets mimic the properties of the real-world data sets and can serve as benchmark data sets.

Despite the outstanding performance of the proposed algorithms, several future research directions are available. Although the \texttt{dQP-SVAS} algorithm is the fastest among the proposed algorithms and provides a competitive solution time for the data sets used in the experiment, it cannot handle a very large data set with millions of subjects. This demands even more scalable algorithms with similar prediction performances. Furthermore, although our study assumes that the subjects are evaluated at the same time, some ratings are from different time periods. Hence, extensions on time series ratings are promising.

\section*{Acknowledgment}
We thank Dr. Y. Zhuang for his technical assistance in collecting a dataset. We also thank the editors and anonymous reviewers for their valuable suggestions and comments that improve the paper.

\section*{Declarations}

\noindent \textbf{Funding} No funding information is available to report.

\noindent \textbf{Conflict of interest} The authors declare that they have no conflict of interest

\noindent \textbf{Code and Data Availability} The real and synthetic data files and codes are attached to the paper as supplements.

\noindent \textbf{Ethics approval} Not applicable.

\noindent \textbf{Consent to participate} Not applicable.

\noindent \textbf{Consent for publication} Not applicable.

\noindent \textbf{Authors contributions} YWP is the lead author of the paper, who initiated the project, collected most of the data sets, developed initial models and algorithms, and performed the experiments. JK developed the estimatability concept and the relevant theorems. YWP and JK improved the algorithms and derived the optimality results. DZ prepared a data set and provided feedback throughout the whole process of the project. All authors are involved in the writing of the paper.

\bibliographystyle{plainnat}
\bibliography{rating_impute}


\pagebreak

\setcounter{figure}{0}
\renewcommand\thefigure{A\arabic{figure}}

\section*{Appendix A. Data Preparation and Description}
\label{section_appendix_data_prep}

Missing rating data frequently appear in practice. In the current research, we are particularly interested in data such that decision-makers or stakeholders might take advantage of imputing the missing entries. For example, customers can use the imputed values to find a service provider that would likely offer better customer service. Institutions can use them to evaluate unrated subjects. Some companies use it to make intelligent recommendation systems.
We compiled multiple real-world data sets in various applications: hospital rating, journal rating, environmental, social, and governance (ESG) rating, elementary and high school rating, and movie rating.
In Sections A1 - A5, we describe the detailed background and our data collection/compiling procedures for each data set. Section A6 presents the conversion procedure of the continuous rating scores. Finally, in Section A7, we present the detailed statistics and distribution of the missing values in the collected data sets.

\subsection*{A1. Journal Data} 
\label{section_data_prep_journal}

The quality of journal publications has been adopted as one of the most significant criteria for faculty scholarship evaluation. The baseline indices are frequently defined based on several important factors, such as the publication volume, citation metrics, acceptance rates, Journal Impact Factor (JIF), H-index, SCImago Journal Rank (SJR) indicator, CiteScores, Source Normalized Impact per Paper (SNIP), and the Eigenfactor, among others.
Despite the existence of such indices, these quality measures are too fragmentary to be adopted as sole quality quantifiers for a journal.
Alternatively, some organizations have developed journal rating systems that determine journal ratings by considering various quality indices along with inputs of expert panels. Such journal rating lists in the business research domain include the Australian Business Deans Council (ABDC), the Association of Business Schools (ABS), the Erasmus Research Institute of Management (EJL), the European Journal of Information Systems (EJIS), and the High Council for Evaluation of Research and Higher Education (HCERES).

Many academic institutions adopt a journal quality list and use the referred ratings as a quality score for a journal publication. Nevertheless, it is commonly observed that some quality journals are not included in the journal list. This necessitates estimating the ratings of the journals within the quality principles of the designated journal list.

In this study, we use the recent journal rating data from Harzing.com \citep{Harzing}, which provides ratings for a collection of RPs. All raw ratings, such as $\{A^*, A, B, C\}$, are converted to numerical values $\{4,3,2,1\}$ such that higher rating values indicate better quality. Note that \cite{kim2019} also use an earlier version of the data from Harzing.com, although the data set differs slightly in terms of the rating values and the number of RPs and subjects. Hence, the results are not directly comparable to the results of the current paper.

\subsection*{A2. Hospital Rating Data}
\label{section_data_prep_hospital}

Hospital administrators, healthcare consumers, and journalists are all captivated by the release of general hospital ratings and their implications for healthcare quality improvement and data analytics. In the United States, these ratings may come from a variety of sources, such as U.S. News and Consumer Reports \citep{ozcan2008health}. Recently, hospital ratings have been publicly released by U.S. government agencies such as the Centers for Medicaid and Medical Services (CMS) and other associated entities under the U.S. Department of Health and Human Services (HHS), sourced from the Agency for Healthcare Research and Quality (AHRQ). Close to 4,000 hospitals in the U.S. are assigned ratings of 1-5 stars \citep{Medicare}. The Leapfrog Group, a U.S. nonprofit watchdog organization, also compiles data for informed purchasing in the healthcare industry to foster hospital transparency and value-based purchasing in the U.S. healthcare system. The evaluations are typically based on hundreds of different measures assessing the hospital facility, services, and equipment quality; a summarized rating is then composed according to the different weights assigned to different categories \citep{venkatesh2018overall}.

To create a combined hospital quality list, we collect ratings from four publicly available rating systems: HCAHPS (Hospital Consumer Assessment of Healthcare Providers and Systems) Star Ratings \citep{Medicare}, LeapFrog's Hospital Safety Grade \citep{LeapFrog}, the Centers for Medicare \& Medicaid Services \citep{Medicare}, and U.S. News's Best Hospital \citep{USNEWS}. For notational convenience, we denote them as \textit{HCAHPS}, \textit{LeapFrog}, \textit{Medicare}, and \textit{U.S. News}, respectively. U.S. News provides ratings of hospitals for 16 specialties and nine procedures and conditions. Among these, we focus on 12 specialties that are not exclusively based on reputation survey and that have numerical scores, following \cite{austin2015national}. The specialty scores are then standardized and averaged. The procedures for the other three sources of rating data (LeapFrog, Medicare, and HCAHPS) are relatively straightforward: A, B, C, D, and F are converted into 5, 4, 3, 2, and 1 for LeapFrog, while the 5-to-1 star rating systems of Medicare (Hospital overall rating) and HCAHPS (Overall hospital rating - star rating) are directly used without modifications. With these ratings from the four rating providers, we create a combined rating list by merging based on the available hospital name, address, city, state, and zip code. The combined rating data include 4,217 hospitals rated by four RPs.

\subsection*{A3. Environmental, Social, and Governance Data}
Environmental, social, and governance (ESG) assessments have become the standards used to measure a firm’s degree of involvement in socially responsible activities \citep{huang2019environmental}. ESG ratings have been widely used in various domains, including socially responsible investing, risk identification \& management, and supplier selection \& evaluation. Given the popularity of ESG measures in both academia and industry, there are several proprietary databases on the market that specialize in ESG ratings for firms.
Firms are evaluated by a score derived from their performance in multiple indicators for each category: this varies across ESG databases. The overall performances can also be evaluated with composite scores.

Among the multiple databases available, we include four databases in our data set: (1) Kinder, Lydenberg, Domini, and Company (KLD), which is now known as MSCI ESG, (2) Bloomberg ESG, (3) Sustainalytics, and (4) CSRHub. KLD \citep{KLD} measures ESG scores based on binary indicators \citep{halbritter2015wages} across three categories (environmental, social, and governance). However, only the binary indicators are reported, while category or composite scores are not reported. To obtain a single score for each category, we follow the common approach that computes the difference between the strengths and concerns \citep{waddock1997corporate}. The overall ESG score is then obtained by aggregating the scores of the three categories. 
The ratings in the Bloomberg ESG data \citep{Bloomberg} primarily indicate the extent to which a firm discloses information regarding ESG. The Bloomberg ESG includes a firm’s overall ESG score and ratings in each of the three categories. The score is updated annually and is tailored based on the industry, with the scale for the scores ranging from 0.1 to 100 \citep{mcbrayer2018does}.
Sustainalytics \citep{Sustainalytics} includes a firm’s ratings in the environmental, social, and governance categories based on multiple indicators. Sustainalytics reports a weighted overall ESG score, where the weights are determined by the firm’s industry. The scale of this overall ESG score is from 0 to 100 \citep{surroca2010corporate}. 
A firm’s overall ESG score in CSRHub \citep{CSRhub} is based on the weighted combinations of the four categories (i.e., community, employees, environment, and governance), with a scale from 0 to 100. CSRHub also reports the overall score every month. We use the simple average of the 12 months as the annual score.  

In this study, we focus on manufacturing firms (NAICS 31-33) in North America in 2014 from COMPUSTAT. Because KLD had stopped reporting the total numbers of strengths and concerns for each category after 2014, which are used to calculate the overall KLD score, the combined rating data use the year 2014. These ESG databases were merged based on the ticker symbol. It should be noted that missing values appear randomly due to the lack of information on the construction of the ESG score.

\subsection*{A4. Elementary and High School Rating Data}
\label{section_data_prep_school}

School rating systems have become widespread for various purposes.
School administrators have adopted school ratings in school accountability reports as a baseline school performance measure to inspect their administrative strategies. This, in turn, provides guidelines for teachers to take action to improve their instructions and testing strategies.
Highly rated schools also use school ratings for advertisements to attract good students.
School quality, typically and simply represented by school ratings, is known to be a significant factor for house prices \citep{Lerner}. Students and parents often use school ratings to choose schools or school districts or provide incentives for school administrators to improve their pedagogical tactics.

Multiple organizations publish school ratings. For example, the Department of Education and its equivalents in state governments in the United States issue school ratings to compare school quality in the state. Some private organizations also issue school ratings for schools across the country. See \cite{dalton2017landscape} for a taxonomy of school rating issuers.

Missingness in school rating systems occurs frequently. In a state school rating system, the missingness results from its geographical constraint. For consumer-oriented rating systems, despite their wide span of subjects, there are missing subjects in the list because each rating issuer maintains its own exclusion rules. For example, Niche.com measures the completeness of data for individual schools, and if a school does not provide a sufficient amount of data, the school is disqualified from the Niche rating.

We prepare two data sets for elementary and high school rating data for Dallas, Texas. To create a combined school rating list, we collect ratings from five publicly available rating systems: GreatSchools \citep{GreatSchool}, Niche \citep{Niche}, SchoolDigger \citep{SchoolDigger}, TexasSchoolGuide \citep{TexasSchoolGuide}, and TexasSchoolRating \citep{TexasSchoolRating}. The original ratings of GreatSchool and SchoolDigger in 10-to-1 and 5-to-0 scales, respectively, are converted into 5-to-1 and 6-to-1 scale ratings. The other RPs provide letter grades (\{A+,…,D-\} and \{A+,…,D-,F\}), which are converted into 4-to-1 and 5-to-1 scale ratings, respectively, based on the letter grade and ignoring the plus or minus. After creating a combined rating matrix by merging based on name and address, we keep the schools that have at least one rating available.

\subsection*{A5. Movie Rating Data}
\label{section_data_movie}
The movie rating data sets have been studied extensively in the collaborative filtering and matrix completion literature and share a similar structure with other data sets presented in this study.
In other words, the users and movies are rating providers and subjects, respectively, in the context of our research.
A movie recommendation system could use the imputed values to build recommended movie lists customized for individual users.

We remark that there are a few differences between movie rating data and other test data. Perhaps, the most striking distinction of the movie rating data from the other experimental data in this paper is its high missing rate. For example, the Movielens data sets \citep{harper2015movielens, MovieLenz} have missing rates ranging from 94.1\% to 99.7\% while the other data sets in Table \ref{table_data_summary} have missing rates from 30.2\% to 55.6\%. Another critical difference is the size of the data. Even the smallest MovieLens data set includes thousands of movies and hundreds of users. Most imputation algorithms cannot handle this data. Lastly, the movie rating data has smaller user correlations than the other data sets we consider. Hence, the proposed imputation methods are unlikely to yield satisfactory imputations with the original form of movie rating data. For these reasons, we customize movie rating data to have similar sizes, missing rates, and inter-rater correlations with the other experimental data.

We use the MovieLens 100k data set \citep{harper2015movielens, MovieLenz} to create a reduced data set with highly correlated rating providers. The original data includes 100,000 ratings from 943 users on 1682 movies, where 1-5 rating system is used for the ratings. To create coherent ratings without too many missing values, we first select the users who rate at least 20\% of the movies. Next, we check the Kendall rank correlations to further reduce the user set by selecting the users whose average correlation is greater than 0.25. Finally, we keep the movies rated at least once by the final user set. The reduced data set with highly correlated users include 12 users and 1102 movies.

\subsection*{A6. Conversion of Continuous Scores}
\label{section_data_prep_conversion}

Our study considers ordinal rating data. When an RP provides continuous scores instead of explicit rating categories, the scores can be converted. Given an RP and scores for all subjects rated by the RP, we partition the subjects into five groups based on their scores and assign 1-5 scale ratings, where 5 is the best and 1 is the worse. Let $d_1,d_2,d_3,$ and $d_4$ be the cutoffs defining five ranges, which are defined below. Given rating provider $j$ for subject $i$ with a continuous score of $s_{ij}$, the converted rating $r_{ij}$ is determined by the following rule:
\[
    r_{ij} = \left\{
\begin{array}{cl}
5 &\textup{if $d_4 \leq s_{ij} < \infty$}\\
4 &\textup{if $d_3 \leq s_{ij} < d_4$}\\
3 &\textup{if $d_2 \leq s_{ij} < d_3$}\\
2 &\textup{if $d_1 \leq s_{ij} < d_2$}\\
1 &\textup{if $-\infty < s_{ij} < d_1$}
\end{array}
\right.
\]
Observe that the values of $d_1 - d_4$ significantly affect the distribution of the converted ratings. To resemble the distribution of the original scores, we use $(d_1,d_2,d_3,d_4) = (0.1,0.35, 0.65, 0.9)$ for US News of the hospital data, which is similar to Normal distribution. For the ESG data, we divide the full range of the scores into equal-length ranges, in order to preserve the distribution of the original scores. For rating provider $j$, we define $s_j^{max}$ and $s_j^{min}$ to be the 99\%tile and 1\%tile of $s_{ij}$. With $s_j^{range} = s_j^{max} - s_j^{min}$, we define $d_j = s_j^{min} + \frac{j*s_j^{range}}{5}$, for $j \in \{1,2,3,4\}$.

\subsection*{A7. Missing Rates by Row and Column}

\begin{figure}[ht]
     \begin{center}
        \subfigure[Hospital]{%
           \includegraphics[scale=0.55]{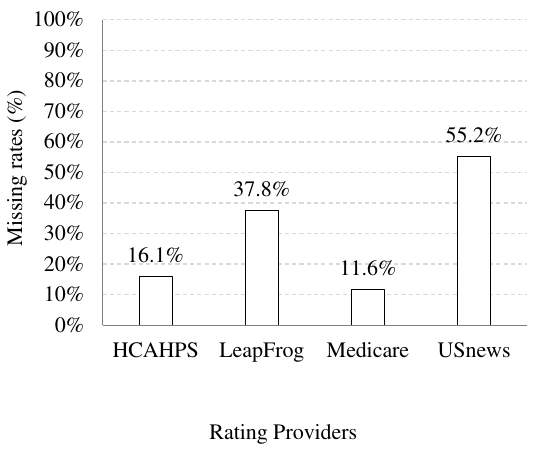} \label{fg_hospitals_missrate}
        }\quad
        \subfigure[Journal]{%
          \includegraphics[scale=0.55]{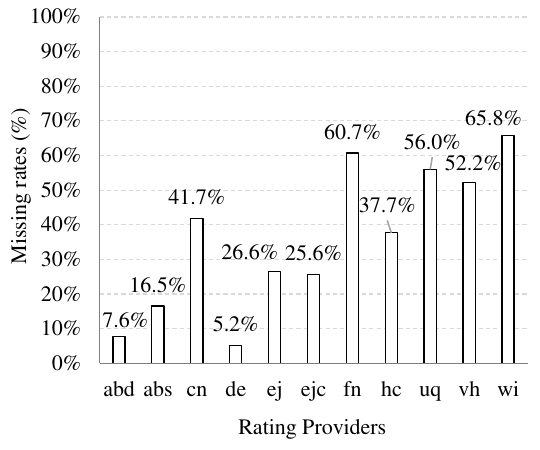} \label{fg_journal_missrate}
        }\quad
        \subfigure[ESG]{%
          \includegraphics[scale=0.55]{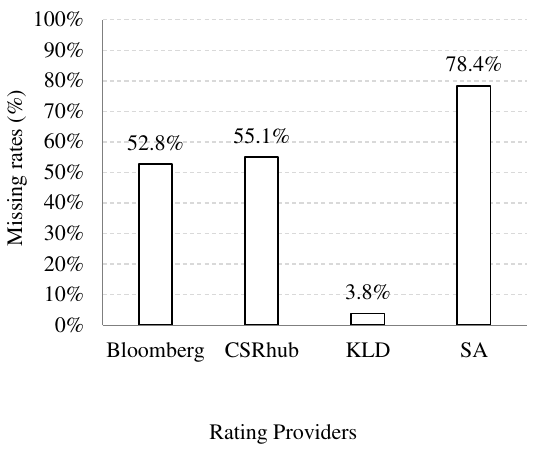} \label{fg_sustainability_missrate}
        }\quad
        \subfigure[Elementary School]{%
          \includegraphics[scale=0.55]{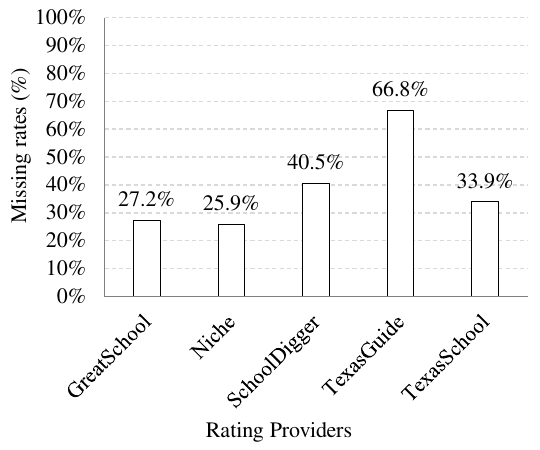} \label{fg_school_elementary_missrate}
        }
        \quad
        \subfigure[High School]{%
          \includegraphics[scale=0.55]{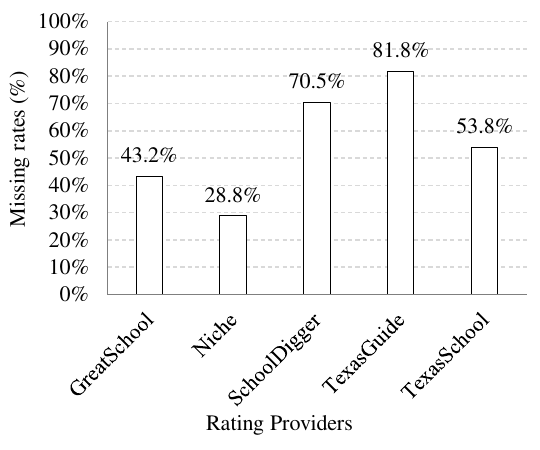} \label{fg_school_high_missrate}
        }\quad
        \subfigure[Movielens]{%
          \includegraphics[scale=0.55]{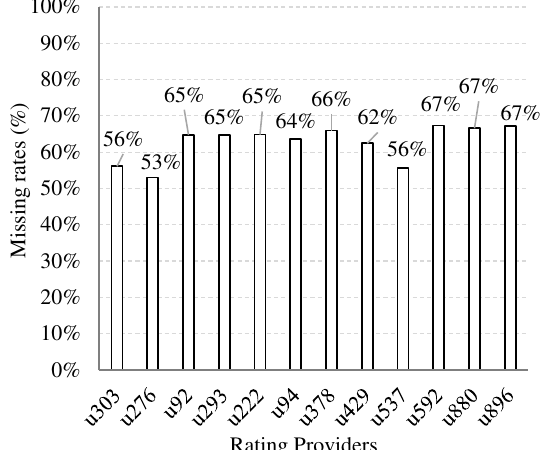} \label{fg_movielens_missrate}
        }
    \end{center}
    \vspace{-0.3cm}
    \caption{Missing rates by rating providers}
\label{fg_missrate}
\end{figure}

\begin{figure}[ht]
     \begin{center}
        \subfigure[Hospital]{%
           \includegraphics[scale=0.55]{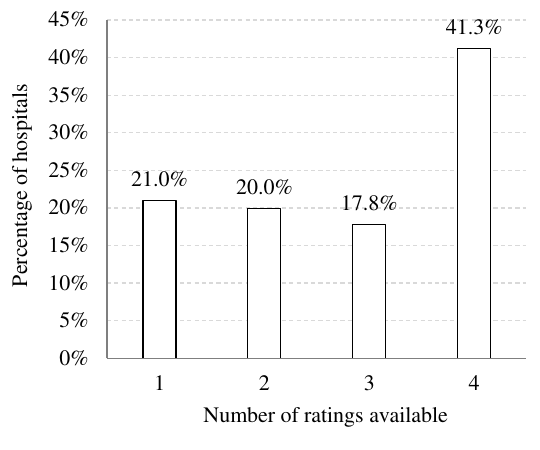} \label{fg_hospitals_num_ratings}
        }\quad
        \subfigure[Journal]{%
          \includegraphics[scale=0.55]{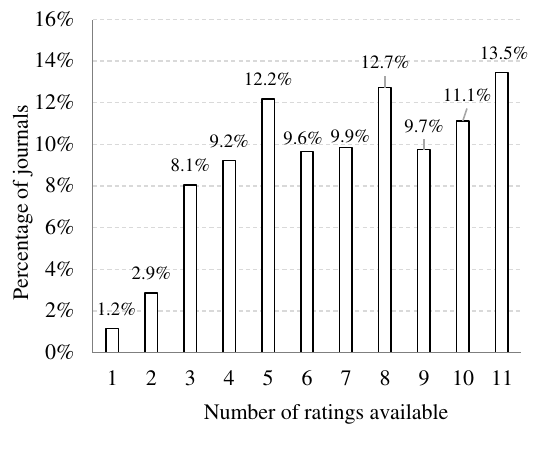} \label{fg_journal_num_ratings}
        }\quad
        \subfigure[ESG]{%
          \includegraphics[scale=0.55]{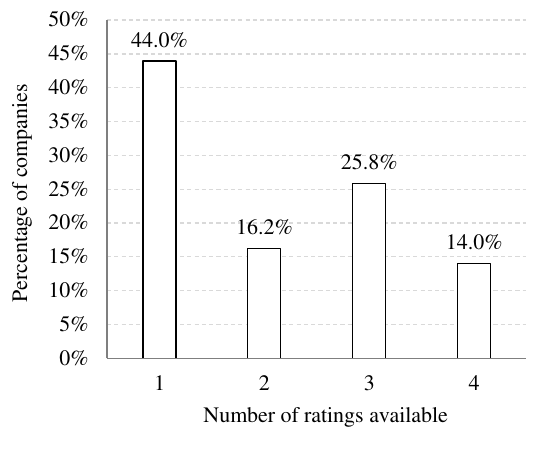} \label{fg_sustainability_num_ratings}
        }\quad
        \subfigure[Elementary School]{%
          \includegraphics[scale=0.55]{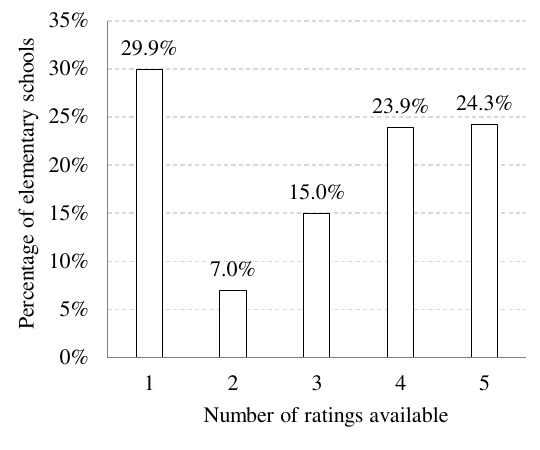} \label{fg_school_elementary_num_ratings}
        }
        \quad
        \subfigure[High School]{%
          \includegraphics[scale=0.55]{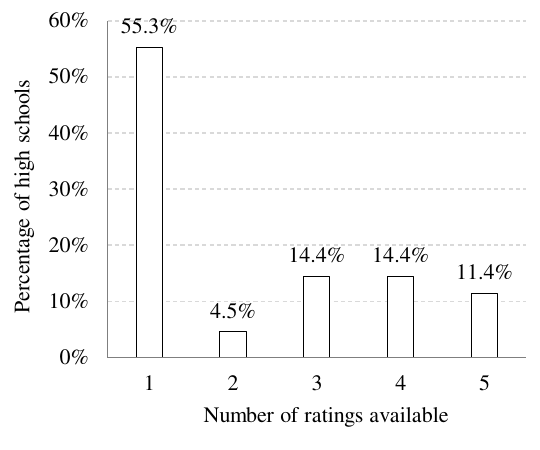} \label{fg_school_high_num_ratings}
        }\quad
        \subfigure[Movielens]{%
          \includegraphics[scale=0.55]{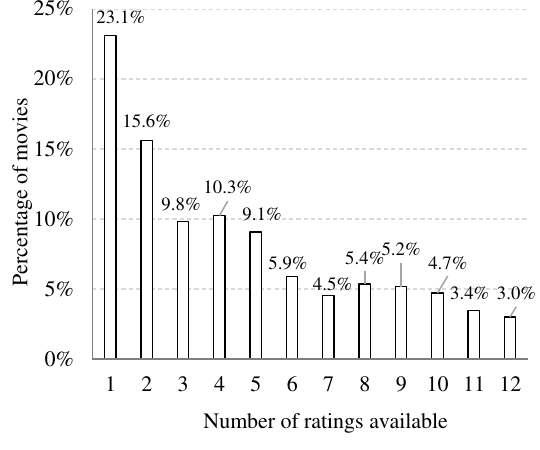} \label{fg_movielens_num_ratings}
        }
    \end{center}
    \vspace{-0.3cm}
    \caption{Available ratings by row}
\label{fg_num_ratings}
\end{figure}

In this section, we present the detailed statistics of missing rates of the data sets. Figures \ref{fg_missrate} and \ref{fg_num_ratings} 
present the missing rates by column and row, respectively, while Table \ref{table_data_summary} in Section \ref{subsection_real_data} only focuses on the summary statistics at the data set level. Hence, this section can help understand the distribution of missing rates over columns and rows.

Figure \ref{fg_missrate} visualizes missing rates for individual rating providers for all data sets, where the horizontal and vertical axes of each plot represent rating providers and missing rates in percentages. Each bar indicates the missing rate of the corresponding RP. For example, 16.1\% for HCAHPS in Figure \ref{fg_hospitals_missrate} indicates that 16.1\% of the 4,217 hospitals have missing ratings by HCAHPS. The missing rates vary across RPs and data sets, from 3.8\% to 81.8\%. Within the same data set, the ESG data set has RPs with missing rate 3.8\% to 78.4\%.

Figure \ref{fg_num_ratings} visualizes the proportions of subjects in percentages by the number of observed ratings. Each bar represents the percentage of subjects for the corresponding number of ratings available. For example, in Figure \ref{fg_hospitals_num_ratings}, 21.0\% and 41.3\% of the 4,217 hospitals are rated by only one RP and all four RPs, respectively. The percentages of fully rated subjects also vary across data sets, from 41.3\% to 11.4\%. Among the data sets, the Hospital and Journal data sets have the increasing number of ratings in the increasing number of available ratings. The other data sets generally have the decreasing number of ratings in the increasing number of available ratings.

\section*{Appendix B. Proofs of Lemmas and Theorems}

\noindent \textbf{Proof of Lemma \ref{lemma:ev_ev1}}\\
Suppose $E_v=\emptyset$. This means that there is no $2\times 2$ submatrix such that one entry is not level-$u$ estimatable for all $u\le v-1$, and each of the three remaining entries is either observed or level-$u$ for some $u\le v-1$.
Because $E_v=\emptyset$, the construction of level-$(v+1)$ estimatable entries must be based on only level-$u$ estimatable entries for $u\le v-1$. Because no desirable $2\times 2$ submatrix exists, thus $E_{v+1}=\emptyset$.
Therefore, if $E_v=\emptyset$ then $E_u=\emptyset$ for all $u\ge v+1$. $\hfill$ $\square$

\vspace{0.5cm}

\noindent \textbf{Proof of Theorem \ref{thm_feasibility_check}}\\
Denote the distinct rating providers in $C$ by $j_1,\dots,j_s$. Then, after relabeling, we can assume that $j_1 - j_2 - \dots - j_s$ forms a connected path.
For any $j=1,\dots,s-1$, because RPs $j$ and $j+1$ are connected, there exists a subject $S$ rated by both $j$ and $j+1$. Therefore, for an arbitrary $S'\in\E(j)$, by the existence of a commonly observed subject $S\in\O(j)\cap\O(j+1)$ and the definition of estimatability, we have $S'\in\E(j+1)$, showing that $\E(j)\subseteq\E(j+1)$. Similarly, the opposite inclusion also holds, showing that $\E(1)=\dots=\E(s)$.
Then, for each $l \in \{1,\dots,s\}$, 
\[
\O(C)=\bigcup_{j=1}^s \O(j)\subseteq\bigcup_{j=1}^s \E(j)=\E(l)
\]
where the first equality holds by the definition of $\O(C)$, 
the inclusion holds because $\O(j)\subseteq\E(j)$ for all $j=1,\dots,s$, and the second equality holds because $\E(1)=\dots=\E(s)$.
Finally, consider an arbitrary entry $(i'', j'')\in\O(C)\times C$. Then, $j''=l$ for some $l\in\{1,\dots,s\}$.
Therefore, $S''\in\O(C)\subseteq \E(l)=\E(j'')$, showing that $S''$ is estimatable. $\hfill$ $\square$

\vspace{0.5cm}

\noindent \textbf{Proof of Theorem \ref{thm_feasibility_check2}}\\
Because the ``if" part directly follows from Theorem~\ref{thm_feasibility_check}, it suffices to show the ``only if" part.
Suppose the graph is disconnected.
Let $C_1,\dots,C_r$ be all the connected components of the graph, where $r\ge 2$.
Consider the $r$ submatrices of the input data, where each submatrix $X_i$ corresponds to the rating providers in $C_i$ and the subjects in $\O(C_i)$ for $i=1,\dots,r$. 
By Theorem~\ref{thm_feasibility_check}, all the entries in $D:=\bigcup_{i=1}^r X_i$ are either observed or estimatable.
Moreover, because $\{C_i\}_{i=1}^r$ are mutually exclusive, so are $\{X_i\}_{i=1}^r$ and $\{\O(C_i)\}_{i=1}^r$, respectively. Moreover, by Assumption \ref{assumption_rate_by_one}, it holds that $O(C_1),\cdots, O(C_r)$ partition the set of all subjects. This, in turn, shows that $O(C_i)\subseteq E(C_i)$ for all $i=1,\dots,r$. Consider data entries outside $D$. We claim that none of these entries are estimatable. 
To prove this by contradiction, consider entry $e\notin D$, which is included in the first estimatable batch outside $D$ in the sequential construction of estimatable entries. 
In other words, the estimatability of $e$ is derived by three corner entries in $D$.
Suppose this entry $e$ is associated with subject $i$ and rating provider $j$.
Let $C$ be the connected component that includes $j$.
Then, there exists a subject $i'\ne i$ and a rating provider $j'\ne j$ such that $(i,j'), (i',j)$, and $(i',j')$ entries are estimatable and included in $D$.
We first show that $j$ and $j'$ are not in the same connected component. Assume $j'\in C$. Because $(i, j')\in D$ and $j'\in C$, it holds that $i \in \O(C)$, implying that $(i,j)\in D$, creating a contradiction.
Therefore, we assume that $j'\in C'$ for some connected component $C'$ other than $C$. Let $X'$ be the submatrix associated with $C'$ and $\O(C')$.
Because $(i', j')\in D$ and $j'\in C'$, it holds that $i'\in \O(C')$. 
Similarly, we can show that $i'\in\O(C)$ because $(i',j)\in D$ and $j\in C$. This contradicts the fact that $\O(C)$ and $\O(C')$ are disjoint. Therefore, $e$ is estimatable.
This means that the sequential construction of estimatable entries is discontinued after the construction of $D$. Therefore, the input data set is unestimatable. $\hfill$ $\square$
\vspace{0.5cm}

\noindent \textbf{Proof of Theorem \ref{theorem_level1}}\\
To prove the ``if" part, assume $I = \bigcup_{j\in N_{j'}} I_j$ for all $j'\in J$ and consider an arbitrary missing entry $(i', j')$. Then, there exists $j\in N_{j'}$ such that $i'\in I_j$ and $j \neq j'$. Because $j\in N_{j'}$, there exists $i\in I$ such that $i \ne i'$ and $i\in I_j \cap I_{j'}$. Hence, for the missing entry $(i', j')$, we have ratings available at $(i', j)$, $(i, j)$, and $(i, j')$. This implies that $(i', j')$ is level-1 estimatable. Applying this procedure for all missing entries in $X$, we can show that $X$ is level-1 estimatable.

We next prove the ``only if" part. Assume that $X$ is level-1 estimatable and consider an arbitrary rating provider $j'$. Then, for each missing entry $(i', j')$, there exists $i''$ and $j''$ such that the ratings are available for entries at $(i'',j')$, $(i',j'')$, and $(i'',j'')$ because of the definition of level-1 estimatability. This implies that $j'' \in N_{j'}$ and $i' \in I_{j''}$. By combining the result for all missing entries in $j'$, we can show that $I = \bigcup_{j\in N_{j'}} I_j$ holds for $j'$. Applying this procedure for all RPs in $J$ completes the proof. $\hfill$ $\square$

\vspace{0.5cm}

\noindent \textbf{Proof of Theorem \ref{thm:strongconvexity}}\\
Consider each term 
\begin{equation}
w_{lj} \Big(\frac{x_{kl}-x_{il}}{c_l} + \frac{x_{ij}-x_{kj}}{c_j} \Big)^2
\label{eq:eachterm}
\end{equation}
of the objective function. Its square part is convex because it is a composition of a convex function ($f(x)=x^2$) and an affine function. Then, \eqref{eq:eachterm} is also convex because it is a scalar multiple of a convex function.
Next, assume that the data set is estimatable. Because input data $X$ includes at least one level-1 estimatable missing entry, there exists a term of the form \eqref{eq:eachterm} such that three ratings are known and one rating is unknown. Because this term is a univariate quadratic function that is concave upward, it is strongly convex. 
Finally, because the sum of a strongly convex function and a convex function is strongly convex, the objective function of \eqref{thm:strongconvexity} is strongly convex. $\hfill$ $\square$

\setcounter{table}{0}
\renewcommand\thetable{C\arabic{table}}

\section*{Appendix C. Analysis on Probability of Estimatability}

In this section, we present an analysis on the estimatability of the data sets generated by Algorithm \ref{algo_synthetic_data}. Based on the graph representation in Section \ref{subsec_data_assumption}, we calculate the probability that a generated data set is estimatable. To achieve this goal, we must check whether or not the corresponding graph is connected. The graph connectivity calculation is based on edge connectivities.

Edge connection probability depends on the missing rate $r$ and the number of subjects (rows) $m$. Following Algorithm \ref{algo_synthetic_data}, let us assume all RPs (columns) have the same missing rates and that the number of missing values in each RP is $\lfloor rm \rfloor$. Then, the edge connection probability, denoted as $P_{edge}$, can be derived as follows:
\[
   P_{edge} = \left\{
\begin{array}{cl}
1 &\textup{if $\lfloor rm \rfloor < \frac{m}{2}$}\\
1 - \frac{\binom{\lfloor rm \rfloor}{2\lfloor rm \rfloor-m}}{\binom{m}{\lfloor rm \rfloor}} & \textup{if $\lfloor rm \rfloor \geq \frac{m}{2}$.}
\end{array}
\right.
\]
The edge connection probabilities $P_{edge}$ are calculated over various $r$ and $m$ in Table \ref{table_appendix_edge_prob}. Note that the edge connection probabilities are reasonably high even if the missing rate $r$ is as high as 0.8.

\begin{table}[htbp]
  \centering
  \begin{small}
       \begin{tabular}{|c|ccccc|}
       \hline
    $r \backslash m$ & 50    & 100   & 500   & 1000  & 5000 \\ \hline
    0.60  & 1.0000 & 1.0000 & 1.0000 & 1.0000 & 1.0000 \\
    0.70  & 0.9986 & 1.0000 & 1.0000 & 1.0000 & 1.0000 \\
    0.80  & 0.9175 & 0.9934 & 1.0000 & 1.0000 & 1.0000 \\
    0.90  & 0.4234 & 0.6695 & 0.9962 & 1.0000 & 1.0000 \\
    0.95  & 0.1727 & 0.2304 & 0.7315 & 0.9280 & 1.0000 \\ \hline
    \end{tabular}%
  \end{small}
    \caption{Edge connection probabilities for various $r$ and $m$}
  \label{table_appendix_edge_prob}%
\end{table}%

The graph connectivity depends on the edge connection probability ($P_{edge}$) and the number of RPs ($n$). However, it is difficult to calculate the probability mathematically given $P_{edge}$ and $n$. On the other hand, it is easy to check if a generated graph is connected or not. Therefore, to approximate the graph connection probability, denoted as $P_{connect}$, we generate 10,000 random graphs for each $(P_{edge},n)$ pair and estimate the probability. For this task, we use the \emph{networkx.is\_connected} function in Python. The simulation result is presented in Table \ref{table_appendix_graph_prob}.

\begin{table}[htbp]
  \centering
  \begin{small}
       \begin{tabular}{|c|cccccc|}
    \hline
    \multicolumn{1}{|l|}{$P_{edge} \backslash{} n$} & 4     & 8     & 12    & 16    & 20    & 50 \\
    \hline
    0.1   & 0.0744  &   0.1147  &   0.2680  &   0.4947  &   0.6911  &   0.9988\\
    0.2   & 0.3278  &   0.6800  &   0.9143  &   0.9803  &   0.9955  &   1.0000\\
    0.3   & 0.6139  &   0.9454  &   0.9956  &   0.9993  &   1.0000  &   1.0000\\
    0.4   & 0.8202  &   0.9932  &   0.9999  &   1.0000  &   1.0000  &   1.0000\\
    0.5   & 0.9353  &   0.9996  &   1.0000  &   1.0000  &   1.0000  &   1.0000\\
    \hline
    \end{tabular}%
  \end{small}
  \caption{Graph connection probability for various $P_{edge}$ and $n$} \label{table_appendix_graph_prob}%
\end{table}%

As expected, $P_{connect}$ rapidly increases as $P_{edge}$ or $n$ increases. Even for a small edge probability of $0.2$, the graph connectivity is over 0.9, which demonstrates that Algorithm \ref{algo_synthetic_data} successfully generates an estimatable data set nine out of ten times. Combining the results in Tables \ref{table_appendix_edge_prob} and \ref{table_appendix_graph_prob}, we conclude that Algorithm \ref{algo_synthetic_data} generates an estimatable data set with high probabilities for most of the realistic parameters for $r, m,$ and $n$. For example, if $r = 0.9$, $m = 50$, and $n = 4$, then $P_{connect} = 0.8619$. This implies that Algorithm \ref{algo_synthetic_data} generates an estimatable data set with a probability greater than 0.8619 if $r \leq 0.9$, $m \geq 50$, and $n \geq 4$. With this high success rate, one can repeat Algorithm \ref{algo_synthetic_data} until an estimatable data set is generated.

\setcounter{table}{0}
\renewcommand\thetable{D\arabic{table}}

\section*{Appendix D. Tests for Robustness of Algorithm}

The proposed algorithms, \texttt{QP-AS} and \texttt{dQP-SVAS}, are deterministic. Hence, given a fixed data set, they always return the same set of imputed values and do not address the uncertainty issues of missing value imputation. To check the robustness of the proposed algorithms over varying subsets of the samples, in this section, we present multiple imputation versions of \texttt{QP-AS} and \texttt{dQP-SVAS}. For each imputation trial, we randomly sample 80\% of the rows (subjects) in the data and run \texttt{QP-AS} (or \texttt{dQP-SVAS}). The iterations end when each of the rows is sampled at least 10 times. We refer to these algorithms as \texttt{QP-ASMI} and \texttt{dQP-SVASMI}.

We remark that each sampled data set should be estimatable. Even when the full data set is estimatable, it is possible to have a sampled data set that is not estimatable. Hence, we need to ensure that all sampled data sets are estimatable before calling \texttt{QP-AS} and \texttt{dQP-SVAS}. For each algorithm and data set, we use the following two performance metrics.
\begin{enumerate}[noitemsep]
    \item \%ZeroSD: the percentage of missing entries that have the same imputed value over all samples
    \item AvgSD: the average standard deviation of the imputed value.
\end{enumerate}
By checking the two metrics, we can measure how consistent the imputed values are (\%ZeroSD) and the variances, if not consistent (AvgSD).

In Table \ref{tab:multiple_imputation_variations}, we check the variations of the imputed values over multiple samples using the real experimental data sets in Table \ref{table_experiment_data}. For each algorithm, \#Samples is the number of samples drawn until the termination criteria are met. The two performance metrics and \#Samples are reported for each data set for the two algorithms \texttt{QP-ASMI} and \texttt{dQP-SVASMI}. Observe that at least 77\% and 73\% of the imputed values are identical for \texttt{QP-ASMI} and \texttt{dQP-SVASMI}, respectively. This implies that the proposed algorithms return the same imputed values for over-70\% of the missing entries. Further, The imputed values' average standard deviations are mostly less than 0.1. The results show that both algorithms return consistent imputed values even if the samples differ. 

\begin{table}[htbp]
  \centering
  \begin{small}
       \begin{tabular}{|c|ccc|ccc|}
       \hline
          & \multicolumn{3}{c|}{QP-ASMI} & \multicolumn{3}{c|}{dQP-SVASMI} \\ \hline
    Data set & \#Samples & \%ZeroSD & AvgSD & \#Samples & \%ZeroSD & AvgSD \\ \hline
    Hospital10F & 21.9  & 97\%  & 0.0125 & 21.3  & 96\%  & 0.0125 \\
    Journal10F & 19.7  & 95\%  & 0.0178 & 20.1  & 95\%  & 0.0204 \\
    ESG10F & 20.5  & 77\%  & 0.0978 & 20.4  & 82\%  & 0.0719 \\
    Elementary10F & 19    & 83\%  & 0.0620 & 19.9  & 79\%  & 0.0776 \\
    Highschool10F & 18.7  & 80\%  & 0.0835 & 18.6  & 73\%  & 0.1029 \\
    Movielens10F & 21.2  & 91\%  & 0.0331 & 19.9  & 88\%  & 0.0453 \\
    \hline
    \end{tabular}%
  \end{small}
    \caption{Variations over samples}
  \label{tab:multiple_imputation_variations}%
\end{table}%

The two algorithms, \texttt{QP-ASMI} and \texttt{dQP-SVASMI}, are also compared against their deterministic versions \texttt{QP-AS} and \texttt{dQP-SVAS} in Table \ref{table_multiple_impute_vs_deterministic}. Similar to the results in the main section, the gray cells and boldfaces are used to indicate the best and within-5\%-gaps, respectively, among all algorithms in Table \ref{tab:result_real}. For example, no algorithms in Table \ref{table_multiple_impute_vs_deterministic} have boldfaces or gray cells for the Journal10F data set, because all four algorithms are not close to the best result (missForest) in Table \ref{tab:result_real}. The result shows that \texttt{QP-AS} and \texttt{dQP-SVAS} are slightly better than \texttt{QP-ASMI} and \texttt{dQP-SVASMI}, while the multiple imputation versions return similar quality solutions in terms of the relative gap from the best. There exists only once case (MAD of Movielens10F data set) where either \texttt{dQP-ASMI} or \texttt{dQP-SVASMI} beats the best solution among all benchmarks including \texttt{QP-AS} and \texttt{dQP-SVAS}. On the other hand, the computation times of \texttt{QP-ASMI} and \texttt{dQP-SVASMI} are much worse. Hence, considering the prediction performances and computation time, we conclude the deterministic versions outperform.

\begin{table}[htbp]
  \centering
  \begin{scriptsize}
        \begin{tabular}{|c|c|c|c|c|c|}
        \hline
    Data set & Measure & QP-AS & dQP-SVAS & QP-ASMI & dQP-SVASMI \\ \hline
    Hospital10F & Time  & 0.8   & 0.8   & 19.8  & 10.9 \\
          & Accuracy & \textbf{0.4240} & \textbf{0.4248} & \textbf{0.4234} & \textbf{0.4238} \\
          & RMSE  & \cellcolor[rgb]{ .851,  .851,  .851}\textbf{0.9559} & \textbf{0.9560} & \textbf{0.9564} & \textbf{0.9561} \\
          & MAD   & \textbf{0.6841} & \cellcolor[rgb]{ .851,  .851,  .851}\textbf{0.6833} & \textbf{0.6848} & \textbf{0.6839} \\ \hline
    Journal10F & Time  & 13.8  & 1.8   & 504.5 & 23.7 \\
          & Accuracy & 0.6801 & 0.6898 & 0.6801 & 0.6898 \\
          & RMSE  & 0.5970 & 0.5886 & 0.5970 & 0.5882 \\
          & MAD   & 0.3318 & 0.3220 & 0.3318 & 0.3218 \\ \hline
    ESG10F & Time  & 0.1   & 0.1   & 4.0   & 1.1 \\
          & Accuracy & \textbf{0.4711} & 0.4564 & \textbf{0.4702} & 0.4600 \\
          & RMSE  & \cellcolor[rgb]{ .851,  .851,  .851}\textbf{0.9637} & 1.0337 & \textbf{0.9642} & 1.0161 \\
          & MAD   & \cellcolor[rgb]{ .851,  .851,  .851}\textbf{0.6498} & 0.6978 & \textbf{0.6507} & 0.6853 \\ \hline
    Elementary10F & Time  & 0.1   & 0.0   & 3.1   & 0.7 \\
          & Accuracy & \textbf{0.6373} & \textbf{0.6361} & \textbf{0.6265} & \textbf{0.6265} \\
          & RMSE  & \textbf{0.6453} & \cellcolor[rgb]{ .851,  .851,  .851}\textbf{0.6441} & \textbf{0.6507} & \textbf{0.6517} \\
          & MAD   & \textbf{0.3819} & \textbf{0.3819} & \textbf{0.3916} & \textbf{0.3916} \\ \hline
    Highschool10F & Time  & 0.0   & 0.0   & 0.7   & 0.2 \\
          & Accuracy & \textbf{0.6591} & \textbf{0.6500} & \textbf{0.6636} & 0.6364 \\
          & RMSE  & \cellcolor[rgb]{ .851,  .851,  .851}\textbf{0.6467} & \textbf{0.6743} & \textbf{0.6512} & 0.6938 \\
          & MAD   & \textbf{0.3682} & 0.3864 & \textbf{0.3682} & 0.4045 \\ \hline
    Movielens10F & Time  & 46.4  & 2.4   & 595.2 & 31.1 \\
          & Accuracy & \cellcolor[rgb]{ .851,  .851,  .851}\textbf{0.4678} & \textbf{0.4668} & \textbf{0.4670} & \textbf{0.4676} \\
          & RMSE  & \textbf{0.8993} & \cellcolor[rgb]{ .851,  .851,  .851}\textbf{0.8968} & \textbf{0.9002} & \textbf{0.8963} \\
          & MAD   & \textbf{0.6204} & \textbf{0.6196} & \textbf{0.6215} & \cellcolor[rgb]{ .851,  .851,  .851}\textbf{0.6187} \\ \hline
    \end{tabular}%
  \end{scriptsize}
    \caption{Performance of QP-ASMI and dQP-SVASMI (boldface = within 5\% relative gap, graycell = best)}
  \label{table_multiple_impute_vs_deterministic}%
\end{table}%

\end{document}